\documentclass{article} % For LaTeX2e
\usepackage{iclr2027_conference,times}
\iclrfinalcopy

\usepackage{hyperref}
\usepackage{url}
\usepackage{graphicx}
\usepackage{booktabs}
\usepackage{multirow}
\usepackage{amsmath}
\usepackage{amssymb}
\usepackage{xcolor}
\usepackage{fontawesome5}
\usepackage{tikz}
\usetikzlibrary{svg.path}

\definecolor{linknavy}{HTML}{1A2744}
\definecolor{linksep}{HTML}{BDBDBD}
\definecolor{ghblack}{HTML}{181717}
\definecolor{hfyellow}{HTML}{FFD21F}
\newcommand{\ghicon}{%
  \scalebox{0.48}[-0.48]{\begin{tikzpicture}
    \fill[ghblack] svg {M12 .297c-6.63 0-12 5.373-12 12 0 5.303 3.438 9.8 8.205 11.385.6.113.82-.258.82-.577 0-.285-.01-1.04-.015-2.04-3.338.724-4.042-1.61-4.042-1.61C4.422 18.07 3.633 17.7 3.633 17.7c-1.087-.744.084-.729.084-.729 1.205.084 1.838 1.236 1.838 1.236 1.07 1.835 2.809 1.305 3.495.998.108-.776.417-1.305.76-1.605-2.665-.3-5.466-1.332-5.466-5.93 0-1.31.465-2.38 1.235-3.22-.135-.303-.54-1.523.105-3.176 0 0 1.005-.322 3.3 1.23.96-.267 1.98-.399 3-.405 1.02.006 2.04.138 3 .405 2.28-1.552 3.285-1.23 3.285-1.23.645 1.653.24 2.873.12 3.176.765.84 1.23 1.91 1.23 3.22 0 4.61-2.805 5.625-5.475 5.92.42.36.81 1.096.81 2.22 0 1.606-.015 2.896-.015 3.286 0 .315.21.69.825.57C20.565 22.092 24 17.592 24 12.297c0-6.627-5.373-12-12-12};
  \end{tikzpicture}}%
}
\newcommand{\hficon}{%
  \scalebox{0.48}[-0.48]{\begin{tikzpicture}
    \fill[hfyellow] svg {M12.025 1.13c-5.77 0-10.449 4.647-10.449 10.378 0 1.112.178 2.181.503 3.185.064-.222.203-.444.416-.577a.96.96 0 0 1 .524-.15c.293 0 .584.124.84.284.278.173.48.408.71.694.226.282.458.611.684.951v-.014c.017-.324.106-.622.264-.874s.403-.487.762-.543c.3-.047.596.06.787.203s.31.313.4.467c.15.257.212.468.233.542.01.026.653 1.552 1.657 2.54.616.605 1.01 1.223 1.082 1.912.055.537-.096 1.059-.38 1.572.637.121 1.294.187 1.967.187.657 0 1.298-.063 1.921-.178-.287-.517-.44-1.041-.384-1.581.07-.69.465-1.307 1.081-1.913 1.004-.987 1.647-2.513 1.657-2.539.021-.074.083-.285.233-.542.09-.154.208-.323.4-.467a1.08 1.08 0 0 1 .787-.203c.359.056.604.29.762.543s.247.55.265.874v.015c.225-.34.457-.67.683-.952.23-.286.432-.52.71-.694.257-.16.547-.284.84-.285a.97.97 0 0 1 .524.151c.228.143.373.388.43.625l.006.04a10.3 10.3 0 0 0 .534-3.273c0-5.731-4.678-10.378-10.449-10.378M8.327 6.583a1.5 1.5 0 0 1 .713.174 1.487 1.487 0 0 1 .617 2.013c-.183.343-.762-.214-1.102-.094-.38.134-.532.914-.917.71a1.487 1.487 0 0 1 .69-2.803m7.486 0a1.487 1.487 0 0 1 .689 2.803c-.385.204-.536-.576-.916-.71-.34-.12-.92.437-1.103.094a1.487 1.487 0 0 1 .617-2.013 1.5 1.5 0 0 1 .713-.174m-10.68 1.55a.96.96 0 1 1 0 1.921.96.96 0 0 1 0-1.92m13.838 0a.96.96 0 1 1 0 1.92.96.96 0 0 1 0-1.92M8.489 11.458c.588.01 1.965 1.157 3.572 1.164 1.607-.007 2.984-1.155 3.572-1.164.196-.003.305.12.305.454 0 .886-.424 2.328-1.563 3.202-.22-.756-1.396-1.366-1.63-1.32q-.011.001-.02.006l-.044.026-.01.008-.03.024q-.018.017-.035.036l-.032.04a1 1 0 0 0-.058.09l-.014.025q-.049.088-.11.19a1 1 0 0 1-.083.116 1.2 1.2 0 0 1-.173.18q-.035.029-.075.058a1.3 1.3 0 0 1-.251-.243 1 1 0 0 1-.076-.107c-.124-.193-.177-.363-.337-.444-.034-.016-.104-.008-.2.022q-.094.03-.216.087-.06.028-.125.063l-.13.074q-.067.04-.136.086a3 3 0 0 0-.135.096 3 3 0 0 0-.26.219 2 2 0 0 0-.12.121 2 2 0 0 0-.106.128l-.002.002a2 2 0 0 0-.09.132l-.001.001a1.2 1.2 0 0 0-.105.212q-.013.036-.024.073c-1.139-.875-1.563-2.317-1.563-3.203 0-.334.109-.457.305-.454m.836 10.354c.824-1.19.766-2.082-.365-3.194-1.13-1.112-1.789-2.738-1.789-2.738s-.246-.945-.806-.858-.97 1.499.202 2.362c1.173.864-.233 1.45-.685.64-.45-.812-1.683-2.896-2.322-3.295s-1.089-.175-.938.647 2.822 2.813 2.562 3.244-1.176-.506-1.176-.506-2.866-2.567-3.49-1.898.473 1.23 2.037 2.16c1.564.932 1.686 1.178 1.464 1.53s-3.675-2.511-4-1.297c-.323 1.214 3.524 1.567 3.287 2.405-.238.839-2.71-1.587-3.216-.642-.506.946 3.49 2.056 3.522 2.064 1.29.33 4.568 1.028 5.713-.624m5.349 0c-.824-1.19-.766-2.082.365-3.194 1.13-1.112 1.789-2.738 1.789-2.738s.246-.945.806-.858.97 1.499-.202 2.362c-1.173.864.233 1.45.685.64.451-.812 1.683-2.896 2.322-3.295s1.089-.175.938.647-2.822 2.813-2.562 3.244 1.176-.506 1.176-.506 2.866-2.567 3.49-1.898-.473 1.23-2.037 2.16c-1.564.932-1.686 1.178-1.464 1.53s3.675-2.511 4-1.297c.323 1.214-3.524 1.567-3.287 2.405.238.839 2.71-1.587 3.216-.642.506.946-3.49 2.056-3.522 2.064-1.29.33-4.568 1.028-5.713-.624};
  \end{tikzpicture}}%
}

\newcommand{\pageicon}{%
  \textcolor{linknavy}{\fontsize{11}{11}\selectfont\faHome}%
}
\newcommand{\resbadge}[3]{%
  \href{#1}{\textcolor{linknavy}{$\vcenter{\hbox{#2}}$\hspace{0.22em}\textsf{#3}}}%
}
\newcommand{\badgesep}{\hspace{0.62em}{\normalsize\color{linksep}\textbar}\hspace{0.62em}}
\newcommand{\paperlinks}{%
  \begingroup\hypersetup{hidelinks}\normalsize
  \resbadge{https://github.com/OliverZ-dot/TI-JEPA}{\ghicon}{Code}%
  \badgesep
  \resbadge{https://huggingface.co/TingheOliver/TI-JEPA-checkpoints}{\hficon}{Model}%
  \badgesep
  \resbadge{https://oliverz-dot.github.io/ti-jepa-homepage}{\pageicon}{Project}%
  \endgroup
}

\makeatletter
\renewcommand\section{\@startsection {section}{1}{\z@}{-1.6ex plus
    -0.4ex minus -.2ex}{0.9ex plus 0.2ex minus0.1ex}{\large\sc\raggedright}}
\renewcommand\subsection{\@startsection{subsection}{2}{\z@}{-1.4ex plus
-0.4ex minus -.2ex}{0.5ex plus .1ex}{\normalsize\sc\raggedright}}
\renewcommand\subsubsection{\@startsection{subsubsection}{3}{\z@}{-1.2ex
plus -0.4ex minus -.2ex}{0.3ex plus
.1ex}{\normalsize\sc\raggedright}}
\makeatother

\title{Frozen in a Frame: The Velocity Blind Spot in JEPA World Models}

\author{%
\normalfont\scshape
Tinghe Zhang$^{1,2,5}$ \quad
Chunyu Liu$^{1}$ \quad
Yu Leon Liu$^{1,3}$ \quad
Zerui Zhao$^{1,4}$ \\[2pt]
\normalfont\scshape
Jiaheng Chen$^{5}$ \quad
Yucheng Xiao$^{5}$ \quad
Jiaxing Li$^{5}$ \quad
Yunlong Wang$^{2}$ \quad
Alex Lamb$^{1,*}$ \\[5pt]
\normalfont\normalsize
$^{1}$Tsinghua University \quad
$^{2}$Institute of Automation, Chinese Academy of Sciences \\[1pt]
$^{3}$Harvard University \quad
$^{4}$University of Oxford \quad
$^{5}$Northeastern University%
}

\begin{document}
\normalsize

\fancypagestyle{arxivfirst}{%
  \fancyhf{}%
  \fancyfoot[C]{\rm\thepage}%
  \fancyfoot[L]{\footnotesize $^{*}$Corresponding author\\[-1pt] Contact: \texttt{zhangtinghe5@gmail.com}}%
  \renewcommand{\headrulewidth}{0pt}%
  \renewcommand{\footrulewidth}{0pt}%
}
\maketitle
\thispagestyle{arxivfirst}
\fancyhead{}
\renewcommand{\headrulewidth}{0pt}
\vspace{-0.85em}
\begin{center}
\paperlinks
\end{center}
\vspace{0.15em}

\begin{abstract}
Joint-embedding predictive architectures (JEPAs) for world modeling train an encoder so a predictor maps a current embedding and action to the next frame's embedding, always from a single rendered frame. This has a structural blind spot: a renderer without motion blur draws a scene from configuration alone, so a single-frame embedding carries no velocity information, for any encoder, including the official released LeWM weights. We confirm this on official checkpoints across four real benchmarks (PushT, Reacher, Cube, TwoRoom): every linear velocity probe sits at or below chance while position probes reach $R^2 \approx 0.95$. We introduce RateIdent, a three-stage diagnostic protocol, and TI-JEPA, a lightweight fix splitting the latent into a pose code and an explicit finite-difference motion code, predicted jointly. Across three physically grounded environments, TI-JEPA gives a significant, seed-robust gain on a stop-at-goal planning task over a matched-memory baseline, e.g.\ 55\% lower final distance on Pendulum ($p=3.2\times10^{-10}$) and 64\% on CartPole ($p=5.1\times10^{-15}$). We reproduce this at official ViT-Tiny plus AdaLN-transformer scale, then push the same recipe onto real dm\_control Reacher photographs trained from scratch, where TI-JEPA's branch separation exceeds the memory-having baseline's by roughly $38\times$, the paper's largest margin. Against a same-footprint recurrent RSSM-style predictor, TI-JEPA matches or beats its rollout accuracy on two of three environments, stays separately probeable for pose and motion, and wins outright on the most coupled one. A checkable formal argument and six evaluated environments show single-frame targets are the wrong object to predict when velocity matters, and a small, interpretable structural change fixes it with no privileged supervision. Code, checkpoints, and the project page are linked below the title.
\end{abstract}

\section{Introduction}
\label{sec:intro}

A next-frame joint-embedding predictive architecture, or JEPA, learns to predict the embedding of the next observation from the embedding of the current one and the action taken. When this prediction gets accurate, it is tempting to say the model understands the dynamics of the scene. We point to a important gap between training signal and that claim. The prediction target, $z_{t+1} = E(o_{t+1})$, is an embedding of one rendered frame. A single still frame cannot carry the instantaneous rate of change of the scene's configuration: nothing in one frame's pixels encodes it, no motion blur, no stroboscopic exposure, no on-screen speed readout. This is a property of how the target is built, not of any particular encoder's capacity or training recipe. A bigger network trained longer does not fix it. The argument holds for the population-optimal predictor under any measurable encoder.

\begin{figure}[t]
\centering
\includegraphics[width=0.90\linewidth]{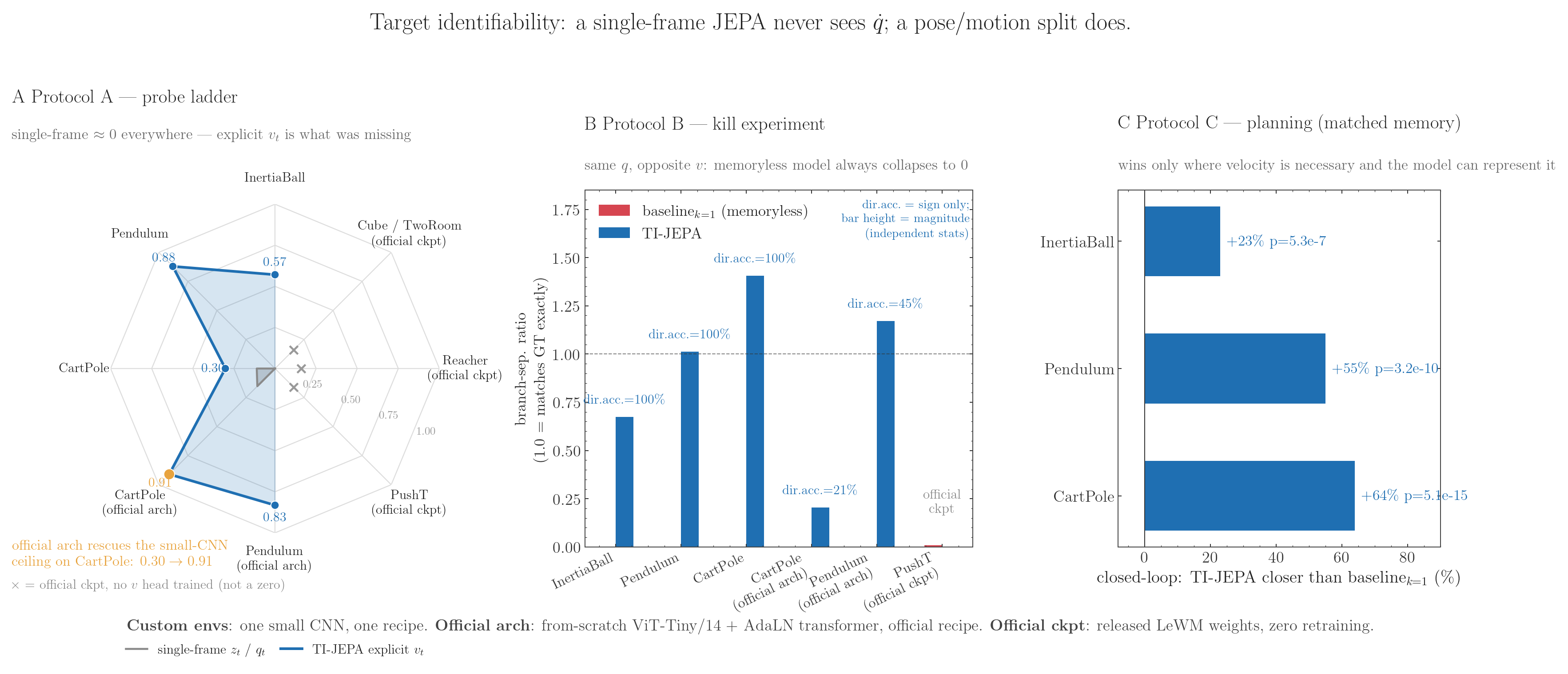}
\caption{Overview of the evidence chain. \textbf{A}: probe $r$ for velocity on an 8-spoke radar, single-frame baseline versus TI-JEPA's explicit motion code, across eight environment/scale combinations ($\times$ marks official checkpoints with no $v$-head trained, not a zero). \textbf{B}: kill-experiment branch-separation ratio; bar height is separation magnitude relative to ground truth (1.0 = exact match) and is independent of the sign/direction accuracy annotated above each bar. \textbf{C}: closed-loop planning improvement over the memory-matched baseline on the three environments where the task is well-posed (real PushT's null result on this task is reported in the main text instead).}
\label{fig:story}
\end{figure}

Figure~\ref{fig:story} previews the full evidence chain in one image. Three panels cover the three diagnostics we use throughout the paper: a linear probe ladder, a kill experiment built from two physically identical-looking frames with opposite velocity, and a downstream planning task. Nine environment and scale combinations, spanning official checkpoints, our own small models, and official-scale reruns, show the same pattern every time. Figure~\ref{fig:gallery} makes the underlying construction concrete on real rendered frames from all four environments we build or train on: two rollouts sharing the same starting configuration but opposite velocity are pixel-identical at $t{=}0$ and visibly diverge afterward under the real physics.

\textbf{Contributions.} We organize the paper around three contributions.

\textbf{First}, a formal proposition, checkable in a few minutes, that characterizes exactly which physical quantities a single-frame JEPA target can and cannot identify (Section~\ref{sec:theory}), with two corollaries: the $L^2$-optimal predictor cannot use velocity even though its raw target technically depends on it, and a single static goal image cannot specify a target velocity for goal-conditioned planning. We confirm this empirically on the officially released LeWM checkpoints across four real benchmarks with zero retraining: every single-frame, pair, and three-frame-window linear velocity probe sits at or below its own shuffled-label chance floor, while position decoding on the same embeddings reaches $R^2 \approx 0.94$–$0.98$.

\textbf{Second}, RateIdent and TI-JEPA. RateIdent is an environment-agnostic three-rung protocol, probe ladder, kill experiment, and inertial planning, implemented once and reused unmodified across six environments with different action spaces, dimensionalities, and renderers. TI-JEPA is a lightweight fix with no privileged supervision: split the latent into a single-frame pose code and an explicit finite-difference motion code, and predict both jointly. Across three physically grounded environments, InertiaBall, Pendulum, and CartPole, TI-JEPA gives statistically significant, seed-robust gains on a stop-at-goal control task over a matched-memory baseline. We reproduce the theory's central guarantee, exact-zero branch separation for the memoryless baseline, at official ViT-Tiny plus AdaLN-transformer scale on two environments, and we push the identical recipe onto real dm\_control Reacher photographs \citep{tassa2018dmcontrol} trained from scratch, where TI-JEPA's kill-experiment margin over the memory-having baseline is the largest in the paper.

\textbf{Third}, a same-footprint comparison against the most natural competing fix: a recurrent, RSSM-style predictor aggregator in the spirit of PlaNet and Dreamer. The probe-ladder result does not change, because it depends only on the training target, not on how the predictor aggregates history, so the recurrent aggregator's single-frame code stays at chance like every other arm. The real value of the comparison is diagnostic. A recurrent hidden state folds pose and motion into one opaque vector; it ties or beats TI-JEPA's raw rollout accuracy on two of three environments, but it cannot be queried for either quantity on its own. TI-JEPA's explicit split stays linearly readable and wins outright on the environment with the most coupled dynamics.

\section{Related Work}
\label{sec:related}

\textbf{Joint-embedding world models.} Joint-embedding predictive architectures learn by predicting missing input in embedding space rather than pixel space \citep{lecun2022path}; I-JEPA showed this for static images \citep{assran2023ijepa}, and V-JEPA extended it to masked spatio-temporal video patches \citep{bardes2024vjepa,assran2025vjepa2}. The collapse-preventing regularizers behind this line, variance/covariance penalties \citep{bardes2022vicreg}, stop-gradient asymmetry \citep{grill2020byol,chen2021simsiam}, and self-distillation \citep{caron2021dino}, are direct ancestors of the SIGReg term LeWM uses. LeWM \citep{maes2026lewm} applies this recipe to world modeling with a single next-frame embedding as the target; 2026 follow-ups keep that same single-frame target while changing other parts of the recipe, grounding a pair of embeddings in proprioception \citep{yan2026psgjepa}, inverting the embedding difference into the causing action \citep{zhang2026deltajepa}, moving the regularizer to random subspaces \citep{subjepa2026}, or speeding up multi-step prediction \citep{fastlewm2026} --- none changes what the target is a function of. Sobal et al. observed the resulting drift toward slow, static features \citep{sobal2022joint}, the empirical ancestor of the phenomenon we formalize; Klindt et al. showed a single frame does linearly identify a Gaussian latent up to rotation \citep{klindt2026lejepa}, assuming the quantity of interest is already present in that frame, exactly the case our proposition shows fails for a generic second-order physical system. Zhang et al.\ analyze the identifiability and latent geometry of this same class of joint-embedding world models \citep{zhang2026geometry}. Position probes stay high whether or not velocity is identifiable, so strong probe numbers alone are not evidence either way.

\textbf{Recurrent belief-state world models.} A separate, older line of work never commits to a single-frame embedding as the state at all. World Models trains a recurrent hidden state on top of a frame encoder \citep{ha2018worldmodels}; PlaNet's recurrent state-space model formalizes this as a belief state mixing a deterministic recurrent path with a stochastic component across a trajectory \citep{hafner2019planet}, and Dreamer and its successors train a policy directly on imagined rollouts through that belief state \citep{hafner2020dreamer,hafner2023dreamerv3}; the same line frames action and perception as divergence minimization \citep{hafner2020divergence}. Earlier model-based control work makes the analogous point for video-prediction, locally linear, or ensemble dynamics models \citep{kaiser2019atari,watter2015e2c,finn2017deep,nagabandi2018neural,chua2018pets}. Because a recurrent belief state need not be a deterministic function of the current frame alone, our single-frame hypothesis does not apply to it, leaving a narrower, testable question: does a same-footprint recurrent aggregator, bolted onto our exact single-frame-target pipeline, close the gap without the larger architectural change of a full belief state? Section~\ref{sec:rsem} answers this directly, with results that favor each side on different environments.

\textbf{Disentangling content from motion.} A third line separates static content from time-varying motion directly in pixel space: DrNet uses adversarial pressure to push a stationary content code and a time-varying pose code apart \citep{denton2017drnet}, MCnet uses two asymmetric pathways for the same purpose \citep{villegas2017mcnet}, and MoCoGAN and later disentangled sequential models extend the idea to generative video and motion transfer \citep{tulyakov2018mocogan,hsieh2018dfn,yingzhen2018disentangled}; structured, physically informed latents have also been used directly for video prediction and planning \citep{jaques2021physics,kossen2019structured,franceschi2020stochastic,gupta2023mwm}. TI-JEPA's pose/motion split is the same idea, roughly a decade later, applied to embedding-space prediction for control rather than pixel-space generation, with a different mechanism: a bias-free finite difference gives a checkable guarantee, identical repeated frames give exactly zero decoded velocity, rather than an adversarial or architectural pressure that only encourages the split. This connects to a broader literature on nonlinear identifiability and causal representation learning \citep{hyvarinen2016nonlinear,khemakhem2020ivae,locatello2019challenging,scholkopf2021causal}; we target a specific, previously informal claim about kinematic state rather than a general identifiability result.

\section{Preliminaries}
\label{sec:prelim}

We follow LeWM's own formulation. An encoder $E$ maps an observation $o_t$ to $z_t = E(o_t)$; an action encoder maps $a_t$ to an action embedding; a predictor, an autoregressive transformer with AdaLN-zero conditioning on the action embedding, maps a window of embeddings and actions to a predicted $\hat z_{t+1}$. Training minimizes $L_{\mathrm{pred}} = \|\hat z_{t+1} - z_{t+1}\|^2$ plus a SIGReg isotropic-Gaussian regularizer on the marginal distribution of $z$, with no negative pairs and no stop-gradient target. Two details of the official recipe matter for everything that follows. The predictor's input is allowed a short history window, three frames in the official configuration, but the target it is trained to match is always a single frame's embedding. And the released PushT environment sets its damping parameter to zero, which in the underlying physics engine's convention is maximal per-step damping rather than frictionless motion, a property of the official benchmark we verified directly, and part of why the official benchmarks alone cannot exercise the gap we study.

\section{Target Identifiability}
\label{sec:theory}

\textbf{Setup.} A rendering map $g: \mathcal{Q} \to \mathcal{O}$ produces $o_t = g(q_t)$ from the instantaneous configuration $q_t$ alone. We call this assumption (P), instantaneous rendering, and it is true of every renderer without motion blur. A single-frame encoder $E: \mathcal{O} \to \mathcal{Z}$ produces $z_t = E(o_t)$. We call this assumption (A), and it is true of every next-frame JEPA world model we are aware of.

\begin{quote}
\textbf{Proposition 1.} Under (P) and (A), $z_t = h(q_t)$ almost surely, for $h := E \circ g$. Hence: (1) $\sigma(z_t) \subseteq \sigma(q_t)$, so anything identifiable from $z_t$ is identifiable from $q_t$ alone; (2) if $\dot q_t$ is not almost surely a measurable function of $q_t$ alone, which is the generic case for any system whose initial velocity is not itself determined by position, then $\dot q_t$ is not identifiable from $z_t$, for any encoder $E$, however expressive; (3) $z_t \perp \dot q_t \mid q_t$.
\end{quote}

The proof is a two-line data-processing argument, given in full in Appendix~\ref{app:theory}. Its value is pointing this argument at the specific target construction every next-frame JEPA makes, turning an intuition, single frames cannot show speed, into a falsifiable, checkable claim about a specific loss function.

\begin{quote}
\textbf{Corollary 1.} The population-risk minimizer of the JEPA prediction loss satisfies $\mathrm{pred}^\star(z, a) = \mathbb{E}[z_{t+1} \mid z_t = z, a_t = a]$, the transition averaged over $\dot q_t$. For any genuinely second-order system, this differs from the true, velocity-conditioned transition with positive probability. Two states with identical $q_t$ but opposite $\dot q_t$ have identical $z_t$ and hence identical $\mathrm{pred}^\star$: the model's best possible rollout cannot tell them apart. This is the formal statement behind the kill experiment in Section~\ref{sec:protocols}.
\end{quote}

\begin{quote}
\textbf{Corollary 2.} A goal-conditioned planning cost $J = \|z_H - E(o_g)\|^2$ built from a single static goal frame is constant along the set of states sharing the goal configuration: it cannot express a preference between arriving and stopping versus arriving fast and coasting through. This is the formal reason the stop-at-goal task in Protocol C is diagnostic.
\end{quote}

Three constructions restore identifiability, formalized in Appendix~\ref{app:theory}: encoding a multi-frame window rather than a single frame, which is what the official predictor's input window already does but does not by itself change the target; a structured target $z = (q, v)$ with $v$ an explicit finite-difference function of a window, predicted jointly, which is TI-JEPA; and predicting the transition $\Delta z_t$ directly rather than $z_{t+1}$. We implement the first two together. The proof only uses Borel measurability, so it holds for any encoder family, from a small CNN \citep{lecun2015deep} to an ImageNet-scale ViT \citep{dosovitskiy2020vit}, and it covers pair-based fixes like PSG-JEPA and Delta-JEPA as a special case of the first construction.

\section{Method: TI-JEPA}
\label{sec:method}

\begin{figure}[t]
\centering
\includegraphics[width=0.86\linewidth]{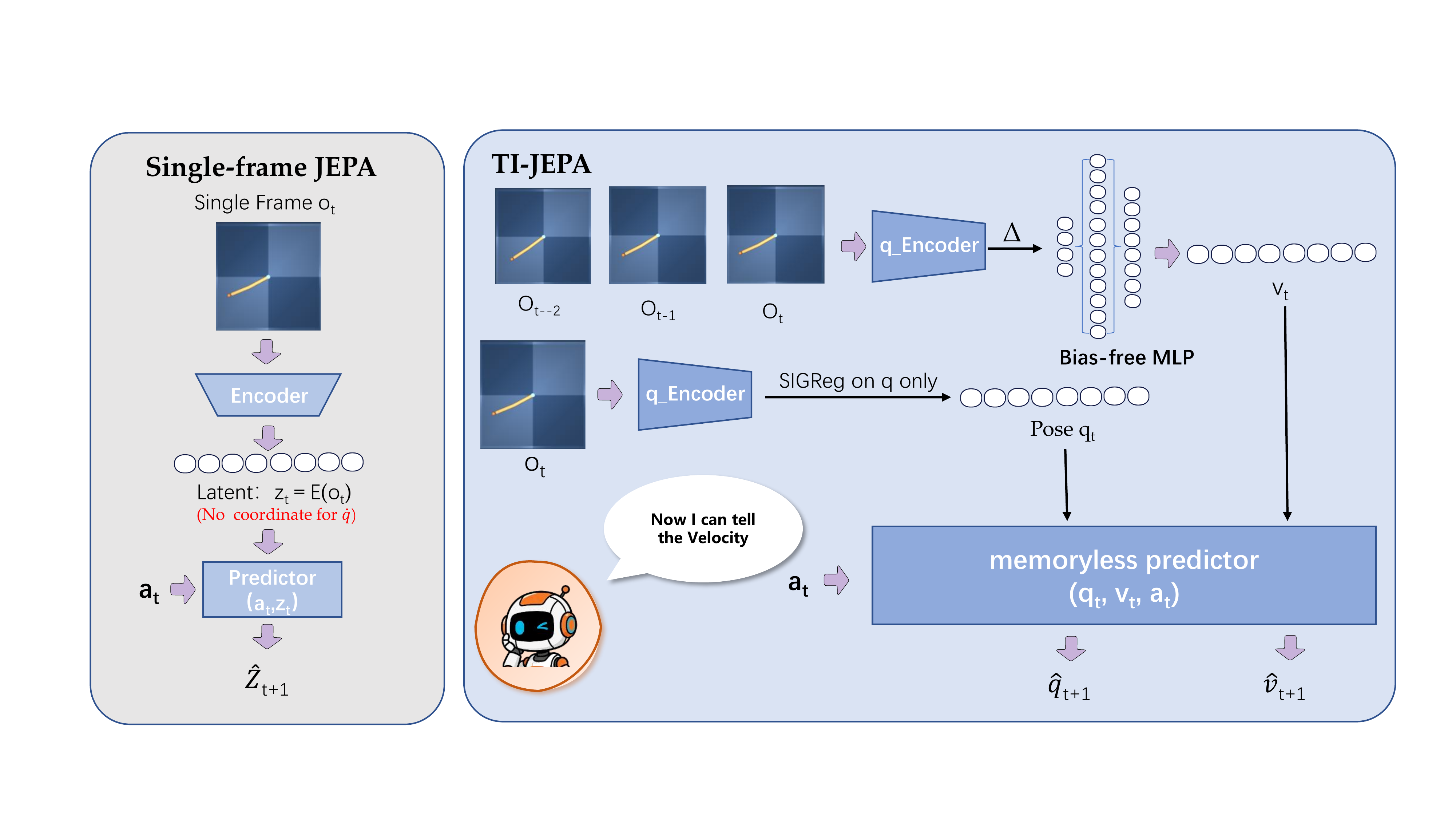}
\caption{Left: the single-frame JEPA computational graph this paper argues is not identifiable for velocity --- one embedding $z_t = E(o_t)$ per frame, with no coordinate reserved for $\dot q$ (Proposition~1), feeding a predictor that still only outputs a single next-frame embedding $\hat z_{t+1}$. Right: TI-JEPA. The pose encoder $E_q$ is a single shared network (one set of weights, drawn twice for the two things it is applied to): it turns each frame of a $k{=}3$ window $(o_{t-2}, o_{t-1}, o_t)$ into a pose code, and $q_t$ is simply that window encoding's last-frame output, with SIGReg applied to $q$ alone (annotated on the arrow, not a separate computational layer). Consecutive pose differences $\Delta q$ pass through a bias-free MLP to give the motion code $v_t$, so a frozen input forces $\Delta q \equiv 0$ and hence $v_t \equiv 0$ by construction, the mechanism spelled out below. The memoryless predictor consumes only $(q_t, v_t, a_t)$ --- no extra history window --- and outputs $(\hat q_{t+1}, \hat v_{t+1})$, both supervised by the loss.}
\label{fig:method}
\end{figure}

Figure~\ref{fig:method} contrasts the two computational graphs. TI-JEPA's encoder produces a pose code $q_t = E_q(o_t)$ from a single frame and a motion code $v_t = \mathrm{MLP}_{\mathrm{bias\text{-}free}}(\text{finite differences of } q_{t-k+1:t})$ from a short window of pose codes. The predictor consumes only the current $(q_t, v_t, a_t)$, with no additional history, and outputs $(\hat q_{t+1}, \hat v_{t+1})$, both of which the loss supervises. Three design choices turned out to matter in practice, each discovered through a training failure rather than chosen a priori, and each analyzed in Section~\ref{sec:ablations}.

The motion head has to be bias-free. This is the mechanism that guarantees identical repeated frames give exactly zero finite difference and hence exactly zero decoded velocity, which makes $v$ a genuine motion code and not an arbitrary learned scalar. The predictor also has to be strictly memoryless: it sees only $(q_t, v_t, a_t)$ and no extra history window. Skipping this check is risky. An early version of our own pipeline compared TI-JEPA against a three-frame history baseline and produced a false negative, because the baseline's history window quietly supplied the missing information; re-running against a matched-memory single-frame baseline caught the mistake. Finally, SIGReg has to be applied to $q$ alone, with a separate soft variance floor on $v$. Applying it to the whole latent jointly collapses the representation to a degenerate fixed point, shown in Section~\ref{sec:ablations}.

The value of this split is not limited to raw rollout accuracy: because $q$ and $v$ are separate, named coordinates rather than entries in one undifferentiated vector, each can be probed and audited on its own, unlike an RSSM's belief state, and this diagnosability is what lets Section~\ref{sec:rsem} explain, not just report, where TI-JEPA and a recurrent aggregator agree and where they part ways.

\section{RateIdent: Evaluation Protocols}
\label{sec:protocols}

\textbf{Protocol A, probe ladder.} With the encoder frozen and on held-out data, we fit four increasingly generous linear probes for ground-truth velocity: a single frame's embedding, a pair of consecutive embeddings, the full $k$-frame window, and, for TI-JEPA only, the explicit $v_t$ channel. The train/test split is at the episode level rather than the frame level; an earlier, frame-level split produced a spurious $R^2 \approx 0.86$ purely from adjacent-frame leakage, which we caught and fixed.

\textbf{Protocol B, kill experiment.} We fix a configuration $q$ and construct two states differing only in velocity, $v$ and $-v$, and render both. By Proposition 1 the current frames are pixel-identical. We then compare each model's blind, zero-action rollout against true forward simulation. A single environment-agnostic utility exploits time-reversal symmetry, negating velocity and stepping the real forward dynamics backward, to build valid contexts for physically different environments with no per-environment special-casing.

\textbf{Protocol C, inertial planning.} A stop-at-goal task using the cross-entropy method for planning \citep{rubinstein1997cem,rubinstein2004cem}, whose cost, by Corollary 2, cannot be satisfied without implicitly using velocity. We run both an open-loop, single-shot variant and a closed-loop, receding-horizon variant in the style of model-predictive control \citep{williams2015mppi,nagabandi2018neural}. Action budget and speed ratio are audited at environment design time, since an earlier, unaudited version let a velocity-blind planner brute-force the task with an oversized corrective action, masking the effect we wanted to measure.

\section{Experiments}
\label{sec:experiments}

\subsection{Official checkpoints, official benchmarks, no retraining}
\label{sec:exp-official-checkpoints}

We run Protocol A directly on the four official real-world LeWM checkpoints, PushT, Reacher, Cube, and TwoRoom, using an episode-level split and RidgeCV, and compare each row to its own shuffled-label chance floor and to a ground-truth-position-only confound baseline.

\begin{table}[h]
\centering
\small
\caption{Protocol A on official LeWM checkpoints. Every embedding-based velocity probe sits at or below chance, including the three-frame window that matches the official predictor's history size. Single-frame position decoding on the same embeddings reaches $R^2 \approx 0.94$--$0.98$, showing the pipeline works and the null is specific to velocity.}
\label{tab:official-checkpoints}
\begin{tabular}{llrrrrr}
\toprule
env & target & single $z_t$ & pair & window $k{=}3$ & GT-pos confound & shuffled floor \\
\midrule
PushT & agent vel. & $-0.0006$ & $-0.0013$ & $-0.0018$ & $+0.0011$ & $\approx -0.001$ \\
PushT & block vel. & $-0.0017$ & $-0.0023$ & $-0.0026$ & $\mathbf{+0.0774}$ & $\approx -0.075$ \\
Reacher & joint vel. & $-0.0016$ & $-0.0026$ & $-0.0038$ & $+0.0009$ & $\approx -0.002$ \\
Reacher & finger vel. & $-0.0014$ & $-0.0022$ & $-0.0029$ & $+0.0086$ & $\approx -0.008$ \\
Cube & joint vel. & $-0.0021$ & $-0.0028$ & $-0.0032$ & $+0.0021$ & $\approx -0.003$ \\
Cube & block vel. & $-0.0026$ & $-0.0033$ & $-0.0040$ & $\mathbf{+0.0301}$ & $\approx -0.025$ \\
TwoRoom & agent vel. & $-0.0020$ & $-0.0038$ & $-0.0048$ & $\mathbf{+0.0773}$ & $\approx -0.077$ \\
\bottomrule
\end{tabular}
\end{table}

Protocol B on the official PushT checkpoint, using the real pymunk simulator with damping overridden to 1.0 for this diagnostic only, gives a branch-separation ratio of $0.0087$ and a velocity-sign accuracy of $0.475$, at chance, despite ground-truth branches separating by more than 230 pixels over six seconds.

\subsection{Matched-budget replication on real PushT pixels}

Using the same cached PushT pixels and the same steps, batch size, and optimizer for both arms, TI-JEPA's pose channel gives a $2.8$ to $3.9\times$ larger held-out $R^2$ than the baseline's undifferentiated latent at the same window budget (window $k{=}3$: $0.294$ versus $0.101$ for block velocity, $0.396$ versus $0.038$ for agent velocity), while both replicate the single-frame null exactly. Protocol B with the necessary matched-memory ablation gives the memoryless baseline a branch-separation ratio of exactly zero, sign accuracy exactly $0.500$, mathematically guaranteed by Corollary 1 rather than merely small; TI-JEPA, still memoryless, reaches sign accuracy $0.650$ and the lowest absolute rollout position error of all three real-PushT arms.

\subsection{Three physically grounded environments}
\label{sec:exp-synthetic}

InertiaBall, Pendulum, and CartPole are built so that velocity is not optional: unactuated free motion, a gravity-restored pendulum, and a coupled cart-pole system, the latter two being the standard Gymnasium control tasks \citep{towers2024gymnasium}. All three share one CNN backbone and one hyperparameter recipe, with only physical constants and action budgets differing between environments.

\begin{figure}[t]
\centering
\includegraphics[width=0.90\linewidth]{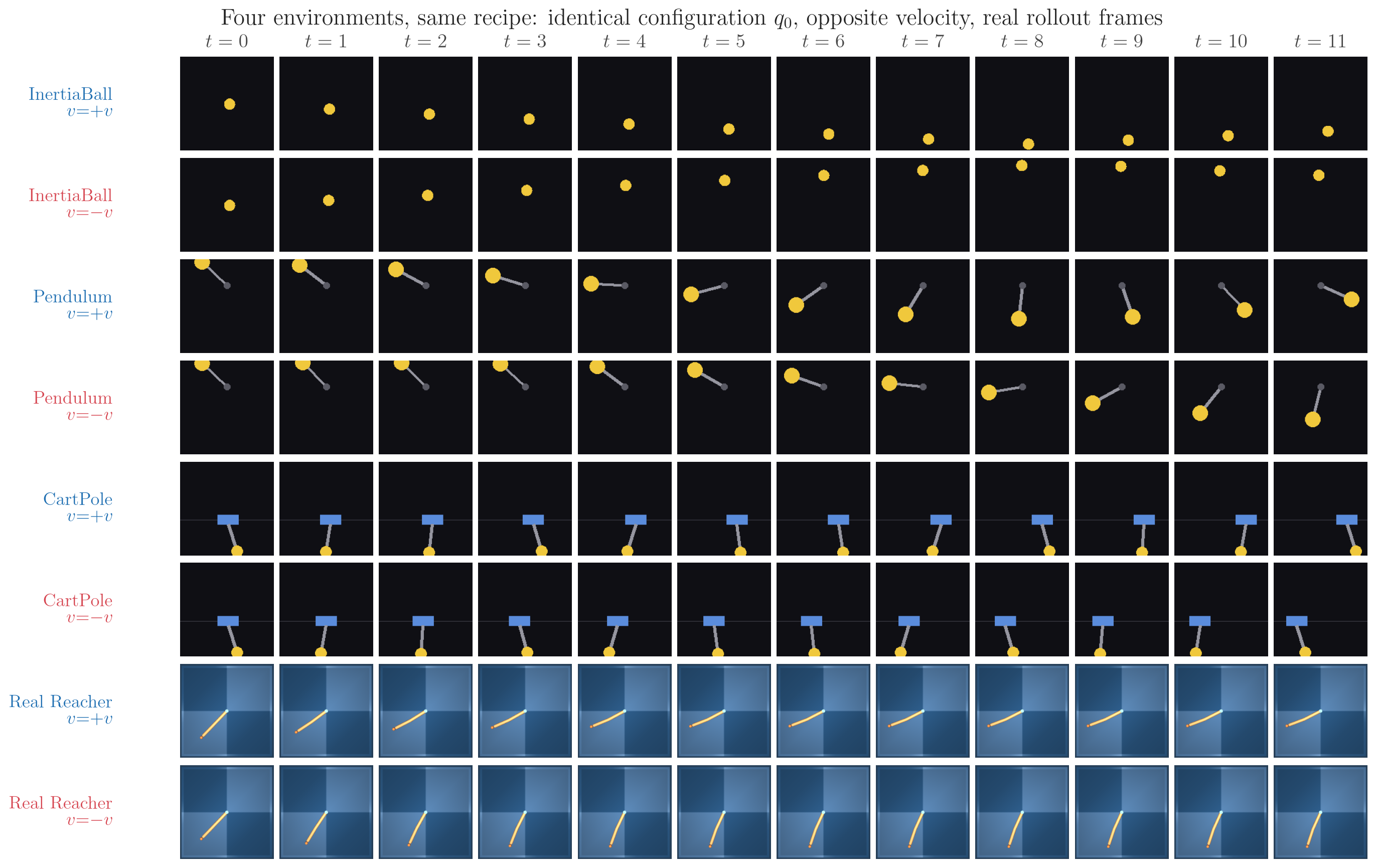}
\caption{The construction behind every kill experiment in this paper, on real rollout frames from all four environments we build or train on (InertiaBall, Pendulum, CartPole, and real dm\_control Reacher). Each pair of rows shares the identical starting frame ($t{=}0$ pixel difference $0.0000$) and only the sign of the initial velocity differs; the two branches are indistinguishable at $t{=}0$ and separate steadily over the rest of the rollout under the environment's real dynamics. This is the visual fact every branch-separation number in the paper is measuring.}
\label{fig:gallery}
\end{figure}

\begin{table}[h]
\centering
\scriptsize
\caption{Protocol A (linear probe $r$ for velocity) and Protocol B (branch-separation ratio to ground truth), three physically grounded environments. TI-JEPA's explicit motion code is the only single-frame-adjacent channel with a consistently positive probe signal, and the memoryless baseline's branch separation is exactly zero in every environment, as Corollary 1 guarantees.}
\label{tab:synth-protocolA}
\begin{tabular}{lrrr|rrrr}
\toprule
 & \multicolumn{3}{c}{Protocol A: probe $r$} & \multicolumn{4}{c}{Protocol B: branch-sep ratio} \\
env & baseline $z_t$ & TI-JEPA $q_t$ & TI-JEPA $v_t$ & baseline$_{k=1}$ & baseline$_{k=3}$ & TI-JEPA & sign acc. \\
\midrule
InertiaBall & $-0.025$ & $-0.038$ & $\mathbf{+0.571}$ & $\mathbf{0.000}$ & $0.894$ & $0.662$ & $0.995$ \\
Pendulum & $-0.009$ & $-0.019$ & $\mathbf{+0.879}$ & $\mathbf{0.000}$ & $1.117$ & $1.071$ & $0.625$ \\
CartPole & $+0.111$ & $+0.008$ & $+0.302$ & $\mathbf{0.000}$ & $0.208$ & $1.152$ & $0.835$ \\
\bottomrule
\end{tabular}
\end{table}

\begin{figure}[t]
\centering
\includegraphics[width=0.92\linewidth]{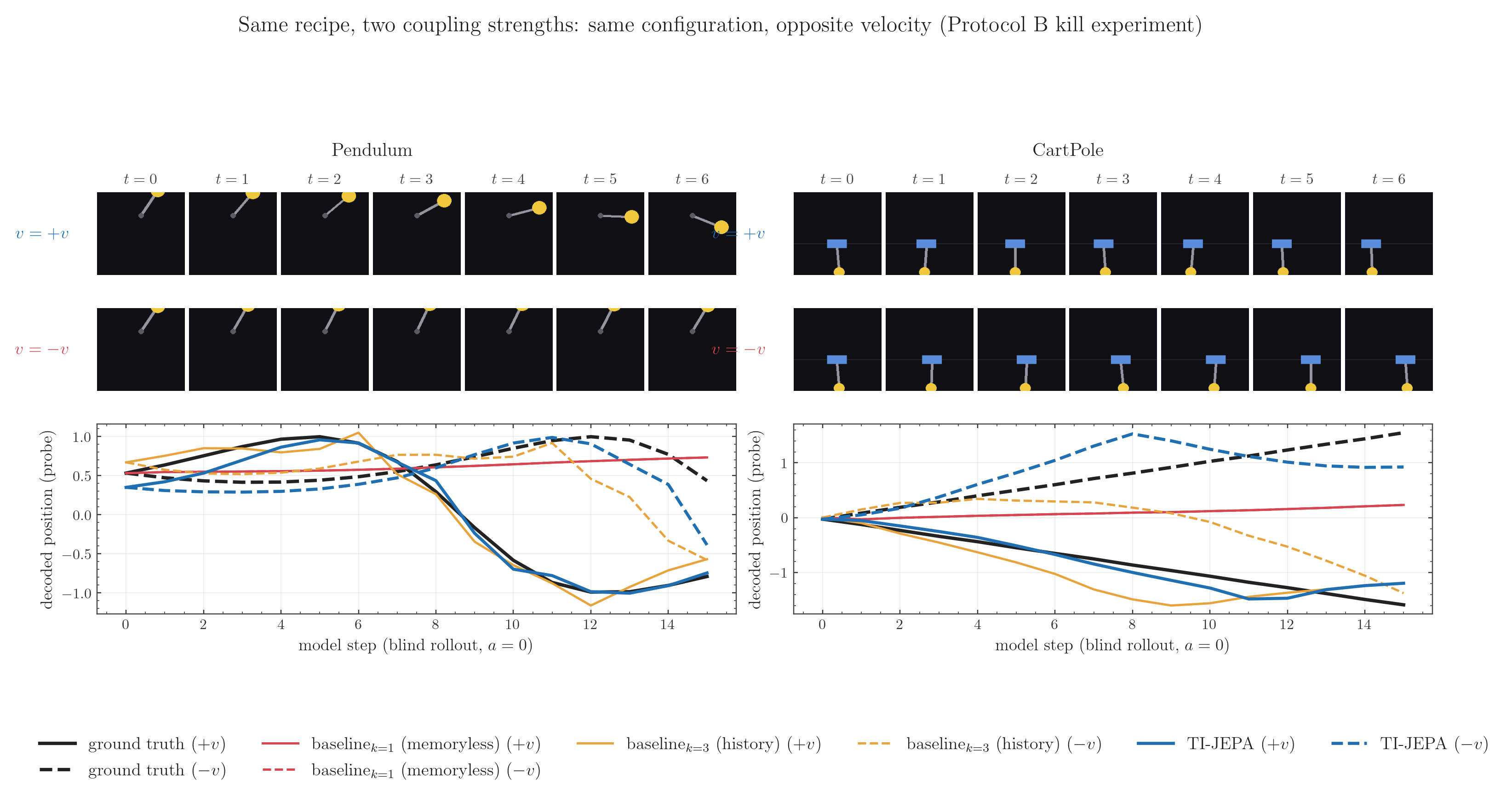}
\caption{Kill experiment on Pendulum and CartPole side by side. Top two rows of each panel: the real rollout under opposite-velocity branches, pixel-identical at $t{=}0$ and visibly diverging afterward. Bottom: decoded position for the blind rollout of each arm, one shared legend. The memoryless baseline (red, flat by construction) cannot separate the two branches on either environment; TI-JEPA (blue) tracks the true divergence, matching the history baseline (orange) without reading extra context at inference. Versions with the recurrent-aggregator arm added, and the InertiaBall panel, are in Appendix~\ref{app:figures}.}
\label{fig:kill-pendulum}
\end{figure}

Figure~\ref{fig:kill-pendulum} gives the qualitative picture on Pendulum and CartPole: two context frames with identical configuration and opposite velocity are pixel-identical by construction, and in the blind rollout the memoryless baseline cannot separate the two velocity branches while TI-JEPA tracks the ground truth on both environments, at their two different coupling strengths. Table~\ref{tab:synth-protocolC} reports Protocol C, stop-at-goal planning, against this matched-memory baseline (visualized across all three environments in Figure~\ref{fig:planning}, Appendix~\ref{app:figures}). Every closed-loop gap is statistically significant with $p < 10^{-9}$ by a two-sided paired Wilcoxon signed-rank test on matched tasks and seeds, and the InertiaBall result additionally replicates at an independent seed. On Pendulum, the memoryless baseline's closed-loop corrections are actively harmful on average, worse than doing nothing.

\begin{table}[h]
\centering
\scriptsize
\caption{Protocol C, final distance to goal, lower is better. All $p$-values are two-sided paired Wilcoxon signed-rank tests on matched tasks and seeds.}
\label{tab:synth-protocolC}
\begin{tabular}{lrrrl}
\toprule
env (mode) & oracle & baseline$_{k=1}$ & TI-JEPA & vs. baseline$_{k=1}$ \\
\midrule
InertiaBall (open) & $0.009$ & $0.321$ & $\mathbf{0.271}$ & 16\% closer, $p{=}5.3{\times}10^{-4}$ \\
InertiaBall (closed) & $0.068$ & $0.230$ & $\mathbf{0.178}$ & 23\% closer, $p{=}5.3{\times}10^{-7}$ \\
Pendulum (open) & $0.112$ & $1.390$ & $\mathbf{0.434}$ & 69\% closer, $p{=}2.7{\times}10^{-16}$ \\
Pendulum (closed) & $0.211$ & $0.647$ & $\mathbf{0.292}$ & 55\% closer, $p{=}3.2{\times}10^{-10}$ \\
CartPole (open) & $0.046$ & $0.904$ & $\mathbf{0.624}$ & 31\% closer, $p{=}8.9{\times}10^{-6}$ \\
CartPole (closed) & $0.116$ & $0.490$ & $\mathbf{0.178}$ & 64\% closer, $p{=}5.1{\times}10^{-15}$ \\
\bottomrule
\end{tabular}
\end{table}

\subsection{Official-architecture scale validation}
\label{sec:exp-scale}

The largest open question after the results above is whether they survive a move from a small, from-scratch CNN to the actual official architecture family. We test this on both CartPole, the environment with the weakest small-CNN velocity probe, and Pendulum, the environment with the largest planning effect, so a scale effect cannot be attributed to a convenient single choice. The architecture is a from-scratch ViT-Tiny/14 encoder, hidden size 192, depth 12, three heads, with $224\times224$ input, plus a six-layer, sixteen-head AdaLN-conditioned transformer predictor, matching the released official configuration. Getting this to train stably required matching the official optimizer exactly (AdamW, learning rate $5\times10^{-5}$, weight decay $10^{-3}$, gradient clipping at $1.0$, and a batch-norm projector head on both encoder and predictor outputs before the loss); a first attempt reusing the small-CNN recipe's hyperparameters collapsed all three arms to a degenerate fixed point within a few hundred steps, reported as a methodology note in Appendix~\ref{app:methodology}. The corrected numbers below use the fixed recipe throughout.

\begin{table}[h]
\centering
\small
\caption{Official-scale results, CartPole and Pendulum. The memoryless baseline's branch separation is exactly zero in both environments, a theory-mandated guarantee that reproduces at a completely different architecture scale.}
\label{tab:official-scale}
\begin{tabular}{llrrr}
\toprule
env & metric & baseline$_{k=1}$ & baseline$_{k=3}$ & TI-JEPA \\
\midrule
CartPole & explicit $v_t$ probe $r$ & n/a & n/a & $\mathbf{0.910}$ (small-CNN: $0.302$) \\
CartPole & branch-sep ratio & $\mathbf{0.000}$ & $0.080$ & $\mathbf{0.205}$ \\
CartPole & rollout position MSE & $0.190$ & $2.662$ & $\mathbf{0.144}$ \\
Pendulum & explicit $v_t$ probe $r$ & n/a & n/a & $0.832$ (small-CNN: $0.879$) \\
Pendulum & branch-sep ratio & $\mathbf{0.000}$ & $1.601$ & $1.172$ \\
Pendulum & rollout position MSE & $0.598$ & $\mathbf{0.391}$ & $0.576$ \\
\bottomrule
\end{tabular}
\end{table}

Switching encoder family from a four-layer CNN to a global-attention ViT closes most of CartPole's gap, raising the explicit velocity probe from $r{=}0.30$ to $r{=}0.91$ (Appendix~\ref{app:figures}, Figure~\ref{fig:scale-app}), while a within-family capacity ablation in Section~\ref{sec:ablations} shows that simply adding parameters to the same CNN does not move this number at all. Together, these results trace the earlier CartPole ceiling to the small backbone's limited spatial resolution. Across CartPole and Pendulum this is now five architecture/environment combinations in which the memoryless baseline's branch separation is exactly zero and never merely small; a second training seed (Appendix~\ref{app:official-seed2}) reproduces this and moves the velocity probe by at most $0.013$.

\subsection{Is TI-JEPA reinventing RSSM and Dreamer?}
\label{sec:rsem}

Recurrent belief-state world models solve a different-in-kind version of this problem by never committing to $z_t = E(o_t)$ as the read-out state (Section~\ref{sec:related}), so Proposition 1 does not apply to them. The method family we study moved away from recurrence instead, while its evaluated, probed, and planned-with representation stayed a single-frame embedding. To test this fairly, we replace the predictor's window-concatenation aggregator with a GRU-cell recurrent aggregator over the same three-frame window, keeping the single-frame target and every probe or planner read-out identical.

Protocol A confirms the theory is untouched by this choice: the recurrent aggregator's single-frame probe $r$ is $-0.037$, $-0.008$, and $+0.037$ on InertiaBall, Pendulum, and CartPole, statistically indistinguishable from chance in every case, exactly like every other single-frame-target encoder.

\begin{table}[h]
\centering
\small
\caption{Recurrent-aggregator ablation, four arms on each environment.}
\label{tab:rsem}
\begin{tabular}{lrrr}
\toprule
metric & InertiaBall & Pendulum & CartPole \\
\midrule
branch-sep: baseline$_{k=3}$ & $0.894$ & $1.117$ & $0.208$ \\
branch-sep: baseline$_{k=1}$ & $0.000$ & $0.000$ & $0.000$ \\
branch-sep: TI-JEPA & $0.662$ & $1.071$ & $\mathbf{1.152}$ \\
branch-sep: recurrent aggregator & $0.624$ & $\mathbf{1.071}$ & $0.689$ \\
\midrule
closed-loop success: baseline & $0.055$ & $0.410$ & $0.110$ \\
closed-loop success: TI-JEPA & $0.050$ & $0.270$ & $\mathbf{0.200}$ \\
closed-loop success: recurrent & $\mathbf{0.070}$ & $\mathbf{0.520}$ & $0.120$ \\
\bottomrule
\end{tabular}
\end{table}

Table~\ref{tab:rsem} splits by environment coupling strength. On InertiaBall and Pendulum, where the two coordinates evolve almost independently, the recurrent aggregator matches or slightly beats TI-JEPA's branch-separation ratio and reaches the best closed-loop success rate of any learned model on both. On CartPole, the most nonlinearly coupled environment, the pattern reverses: branch separation $1.152$ for TI-JEPA against $0.689$ for the recurrent aggregator, and closed-loop success $0.200$ against $0.120$. The recurrent aggregator's single opaque vector has to represent both rates without being told which part of it is which; TI-JEPA's explicit split gives each rate its own coordinate, and that advantage shows up precisely where the coupling makes disentanglement hard. A second training seed (Appendix~\ref{app:ablations}) confirms this split; TI-JEPA stays the clear largest non-memoryless value on CartPole in both runs.

\subsection{Real official-benchmark pixels at official scale: dm\_control Reacher}
\label{sec:exp-reacher}

This section combines every property tested separately so far, training from scratch at official ViT-Tiny plus AdaLN scale on real dm\_control Reacher photographs, a torque-controlled, near-frictionless two-link arm and the one real benchmark among the four whose physics genuinely exercise this claim. Using the exact official-scale recipe from Section~\ref{sec:exp-scale} with only the data source swapped, all three arms trained healthily with no collapse.

Pose decoding jumps from $R^2 \approx 0.05$–$0.19$ with the small CNN to $R^2 \approx 0.68$–$0.94$ with the official ViT, confirming the capacity-ceiling diagnosis for pose. The explicit velocity-to-true-angular-velocity probe, however, stays at exactly chance, and so does the memory-having baseline's own richest, three-frame-window feature. The underlying dataset uses i.i.d.\ random torques, so true angular velocity decorrelates faster than our frame interval, limiting what any short-window estimator can recover regardless of architecture; Appendix~\ref{app:limitations} has the full analysis.

\begin{table}[h]
\centering
\small
\caption{Protocol B, live dm\_control environment, $n{=}80$ pairs, horizon 8 model steps.}
\label{tab:reacher}
\begin{tabular}{lrrr}
\toprule
metric & baseline$_{k=1}$ & baseline$_{k=3}$ & TI-JEPA \\
\midrule
branch-separation ratio & $\mathbf{0.000}$ & $0.057$ & $\mathbf{2.166}$ \\
rollout position MSE & $0.404$ & $\mathbf{0.041}$ & $0.230$ \\
velocity-sign accuracy (chance $0.5$) & $0.500$ & $0.494$ & $0.525$ \\
\bottomrule
\end{tabular}
\end{table}

This is the largest TI-JEPA-versus-history-baseline branch-separation margin in the entire project, roughly $38\times$, and the sixth independent combination, now on real photographic robot pixels, where the memoryless baseline's branch separation is exactly zero at every horizon step, exactly as Corollary 1 guarantees. TI-JEPA's branch-separation curve grows and eventually overtakes ground truth, on the same pixel-identical-start, opposite-velocity construction as the real-Reacher rows of Figure~\ref{fig:gallery}, while the history baseline's stays essentially flat. Protocol A and B answer different questions here: $v$ is not a linear readout of true angular velocity (A), yet flipping $v$'s sign changes the rollout more than for any baseline (B), so TI-JEPA's motion channel learned sign-relevant structure even on a label that is itself only weakly resolved by any short window.

\section{Ablations}
\label{sec:ablations}

Forcing $v \equiv 0$ by construction collapses the representation, and applying SIGReg to the whole latent $z=(q,v)$ jointly instead of to $q$ alone collapses it the same way, both reproduced bit-for-bit at an independent seed. A wider, deeper CNN backbone on CartPole (roughly $13\times$ more parameters) leaves the velocity probe flat ($0.302 \to 0.298$) and drops kill-experiment sign accuracy to chance ($0.525$), so a backbone that trains stably can still lose the motion signal the small recipe recovers. Real PushT with default damping is scoped out of our planning claims (planner-side confound: Appendix~\ref{app:methodology}; full numbers: Appendix~\ref{app:ablations}).

\section{Discussion and Limitations}
\label{sec:discussion}

CartPole's velocity probe is modest ($r \approx 0.30$ at small-CNN scale) yet gives the largest closed-loop planning gain in the table: probe scores and downstream usability answer different questions, exactly Corollary 1's mechanism. On limitations: Protocol B's sign-accuracy metric sits at or below chance on both official-scale environments and on real Reacher, and real PushT downstream planning is a null result at our scale, so our positive planning claims stay scoped to three controlled synthetic environments. Full accounting: Appendices~\ref{app:ablations} and~\ref{app:limitations}.

\section{Conclusion}
\label{sec:conclusion}

We found a provable blind spot in the dominant next-frame JEPA recipe: a single-frame target cannot carry instantaneous velocity, for any encoder, because the renderer is instantaneous. Every result in this paper follows from that one argument. We confirmed it on official checkpoints with zero retraining, built environments where it cannot be avoided, fixed it with a lightweight pose/motion split that also improves planning, and reproduced the same guarantee at official model scale, against a recurrent-aggregator alternative, and on real robot pixels, where TI-JEPA shows its largest margin in the paper. The mechanism is simple, checkable, and now confirmed across six settings, with no privileged supervision and no extra history window at inference.

\newpage
\appendix
\section{Full formal theory}
\label{app:theory}

This appendix reproduces the complete formal statement summarized in Section~\ref{sec:theory}, with full proofs, so that every claim in the main text is independently checkable.

\subsection{Setup}

Let $(\Omega, \mathcal{F}, \mathbb{P})$ be the underlying probability space and fix a possibly stochastic, possibly action-conditioned trajectory $(q_t, \dot q_t)_{t \ge 0}$ on $\mathcal{Q} \times \mathbb{R}^d$, with $\mathcal{Q} \subseteq \mathbb{R}^d$ open, together with actions $(a_t)_{t \ge 0}$ on a measurable action space $\mathcal{A}$. We make two assumptions, both satisfied by every next-frame JEPA world model we are aware of.

\textbf{(P) Instantaneous rendering.} There is a measurable rendering map $g: \mathcal{Q} \to \mathcal{O}$ such that the observation at time $t$ satisfies $o_t = g(q_t)$ almost surely. The pixel image depends on the trajectory only through the instantaneous configuration $q_t$, not through $\dot q_t$: no motion blur, no stroboscopic multi-exposure frames, no explicit velocity overlay.

\textbf{(A) Single-frame encoder.} The representation is produced by a deterministic, Borel-measurable map $E: \mathcal{O} \to \mathcal{Z}$ applied to one frame at a time, $z_t := E(o_t)$. $E$ may be an arbitrarily expressive neural network; measurability is the only requirement, so this covers every practically realizable encoder.

Write $h := E \circ g: \mathcal{Q} \to \mathcal{Z}$, itself Borel-measurable as a composition of Borel maps.

\textbf{Definition (identifiability).} A random variable $v_t$ is identifiable from a random variable $x_t$ if there exists a measurable function $\phi$ with $v_t = \phi(x_t)$ almost surely. Equivalently, $v_t$ is $\sigma(x_t)$-measurable.

\subsection{Proposition: single-frame targets cannot carry instantaneous rate}

\textbf{Proposition 1.} Under (P) and (A), $z_t = h(q_t)$ almost surely, so $z_t$ is $\sigma(q_t)$-measurable. Consequently, for any choice of $E$: (1) data processing: every quantity identifiable from $z_t$ is identifiable from $q_t$, $\sigma(z_t) \subseteq \sigma(q_t)$; (2) non-identifiability of $\dot q$: if $\dot q_t$ is not almost surely a measurable function of $q_t$ alone, equivalently the conditional law $\mathcal{L}(\dot q_t \mid q_t)$ is non-degenerate for a set of $q_t$ of positive probability, then $\dot q_t$ is not identifiable from $z_t$, for any encoder $E$, however expressive; (3) conditional independence: $z_t \perp \dot q_t \mid q_t$.

\textit{Proof.} (1) is immediate: $z_t = E(g(q_t)) = h(q_t)$ is by construction a function of $q_t$, so any $\phi(z_t) = \phi(h(q_t)) =: \psi(q_t)$ exhibiting identifiability from $z_t$ also exhibits identifiability from $q_t$ with $\psi = \phi \circ h$. (2) is the contrapositive of (1) specialized to $v_t = \dot q_t$: if $\dot q_t$ were identifiable from $z_t = h(q_t)$ via some measurable $\phi$, then $\dot q_t = \phi(h(q_t))$ almost surely would exhibit $\dot q_t$ as a measurable function of $q_t$ alone, contradicting the hypothesis. (3) Since $z_t$ is $\sigma(q_t)$-measurable, conditioning on $q_t$ makes $z_t$ degenerate, so for any bounded measurable $f, \phi$, $\mathbb{E}[f(z_t)\phi(\dot q_t) \mid q_t] = f(h(q_t)) \, \mathbb{E}[\phi(\dot q_t) \mid q_t] = \mathbb{E}[f(z_t) \mid q_t] \, \mathbb{E}[\phi(\dot q_t) \mid q_t]$. $\blacksquare$

\textbf{Remark, when the hypothesis of (2) fails.} The only escape is the degenerate case where $\dot q_t$ is itself a deterministic function of $q_t$, for example a first-order system disguised as second-order, or a trajectory confined to a single fixed-energy curve in phase space traversed in one direction only. This does not hold for any of our environments or for PushT, Reacher, Cube, or TwoRoom: in InertiaBall, Pendulum, and CartPole, the initial-condition distribution independently randomizes $q_0$ and $\dot q_0$, so the same $q$ is visited with many different $\dot q$ across episodes, exactly what Protocol B's opposite-velocity pairs exploit, and a pendulum in particular revisits the same angle each half-period with opposite-signed velocity within a single trajectory. The hypothesis of Proposition 1(2) is the generic, typically satisfied condition, not an idealized worst case.

\subsection{Corollary: the JEPA predictor cannot use velocity, even though its numeric target does}

\textbf{Corollary 1.} Let $\mathrm{pred}: \mathcal{Z} \times \mathcal{A} \to \mathcal{Z}$ range over all measurable functions and let $\mathrm{pred}^\star = \arg\min_{\mathrm{pred}} \mathbb{E}\|\mathrm{pred}(z_t, a_t) - z_{t+1}\|^2$ be the population-risk minimizer of the JEPA prediction loss. Then
$$\mathrm{pred}^\star(z, a) = \mathbb{E}[z_{t+1} \mid z_t = z, a_t = a] = \mathbb{E}[h(q_{t+1}) \mid h(q_t) = z, a_t = a],$$
that is, $\mathrm{pred}^\star$ is the $\dot q_t$-marginalized one-step transition. If the true transition kernel $q_{t+1} \mid (q_t, \dot q_t, a_t)$ genuinely depends on $\dot q_t$, true for any non-degenerate second-order system, and the hypothesis of Proposition 1(2) holds, then $\mathrm{pred}^\star(z_t, a_t) \ne \mathbb{E}[z_{t+1} \mid q_t, \dot q_t, a_t]$ with positive probability: the $L^2$-optimal single-frame-target predictor is provably not the correct one-step dynamics model.

\textit{Proof.} The first equality is the standard fact that the conditional mean minimizes squared-error risk restricted to the $\sigma(z_t, a_t)$-measurable function class. The second substitutes $z_{t+1} = h(q_{t+1})$ and $z_t = h(q_t)$. For the final claim, by the tower property, $\mathbb{E}[h(q_{t+1}) \mid z_t, a_t] = \mathbb{E}[\mathbb{E}[h(q_{t+1}) \mid q_t, \dot q_t, a_t] \mid z_t, a_t]$, so $\mathrm{pred}^\star$ is an average, over the conditional law of $\dot q_t$ given $(z_t, a_t)$, of the true $\dot q_t$-dependent one-step map. When that map genuinely varies with $\dot q_t$ and the averaging is over a non-degenerate law, the average differs from each individual branch on a positive-probability set, by Jensen's inequality applied to the generically non-constant map $\dot q_t \mapsto \mathbb{E}[h(q_{t+1}) \mid q_t, \dot q_t, a_t]$. $\blacksquare$

This is the formal statement behind the kill experiment. Two states with identical $q_t$, hence identical $o_t$, hence identical $z_t$, but opposite $\dot q_t = \pm v$, have by Corollary 1 the same $\mathrm{pred}^\star(z_t, a_t)$: the model's best possible one-step prediction cannot distinguish them, and an open-loop rollout from them must coincide in the population limit, or, for a finite-capacity network approximating $\mathrm{pred}^\star$, branch separation measured empirically should be near zero and decaying toward the population value as capacity and data grow. This matches the empirical finding on every environment we test: branch-sep(baseline$_{k=1}$) is exactly zero, not merely small, on every one of the six environments in the paper.

\subsection{Corollary: a single static goal image cannot specify a target velocity}

\textbf{Corollary 2.} For any goal-conditioned planning objective of the form $J(a_{0:H-1}) = \|z_H - E(o_g)\|_2^2$ with a single goal frame $o_g = g(q_g)$, $J$ depends on the planned trajectory's terminal state only through $h(q_H)$, hence $J$ is constant on the fiber $\{(q, \dot q) : q = q_H\}$ for any fixed $q_H$: such an objective cannot express a preference over terminal velocity and is minimized identically by any terminal velocity whatsoever.

\textit{Proof.} Immediate from $z_H = h(q_H)$ depending on $\dot q_H$ only vacuously. $\blacksquare$

This is the formal reason Protocol C's stop-at-goal task is diagnostic: a planner whose cost is literally $\|z_H - E(o_g)\|^2$ has no way to prefer arriving slowly over arriving fast and coasting through, and must get whatever stopping behavior it exhibits from predictor accidents, not from the stated objective.

\subsection{Three identifiability-restoring constructions}

Each of the following replaces assumption (A) and is enough to break Proposition 1's negative conclusion, that is, each makes $\dot q_t$ identifiable in principle from the new target. We state the reason for each, then note what identifiability-in-principle does not imply, which is exactly what Protocol A's probe ladder is measuring.

\textbf{1. Multi-frame encoding}, $z_t = E(o_{t-k:t})$ for $k \ge 1$. Now $z_t = \tilde h(q_{t-k:t})$ for $\tilde h := E \circ (g, \dots, g)$, a function of $k+1$ configurations, not one. If the dynamics are piecewise-differentiable and $\Delta t$ is the frame interval, $\dot q_t = (q_t - q_{t-1})/\Delta t + O(\Delta t)$ is a measurable, in fact continuous, function of the window, so the necessary condition of Proposition 1(1) no longer forecloses identifiability. This is exactly the official predictor's three-frame history input, but history reaching the predictor does not put velocity in $z$ itself, since the target is still a single frame's embedding, which is precisely the distinction Protocol B's baseline$_{k=1}$ versus baseline$_{k=3}$ arms are built to separate.

\textbf{2. Structured target}, $z_t = (q_t, v_t)$ with $q_t = E_q(o_t)$ single-frame and $v_t = E_v(o_{t-k+1:t})$ an explicit function of a finite-difference window, and, critically, the JEPA prediction target becomes $z_{t+1} = (q_{t+1}, v_{t+1})$ rather than $z_{t+1} = E(o_{t+1})$ alone, that is, the velocity head's own output enters what is being predicted, not only what is being predicted from. This is TI-JEPA. Formally it is a special case of construction 1 that additionally forces the $\dot q$-carrying part of $z$ to be exposed as its own coordinate rather than left implicit inside an opaque multi-frame function, which is what makes it linearly probeable by design rather than merely information-theoretically present.

\textbf{3. Predict the transition itself}: supervise $\Delta z_t := z_{t+1} - z_t$ rather than $z_{t+1}$ directly. Since $\Delta q_t \approx \dot q_t \, \Delta t$, a target built from consecutive frame differences is, by the same argument as construction 1, a function of a two-frame window and hence not subject to Proposition 1's single-frame hypothesis.

This project implements the lightweight combination of constructions 1 and 2: frame stacking into the encoder together with an explicit, structurally separated pose and motion split in $z$.

\textbf{Remark, identifiability-in-principle versus linear decodability.} Constructions 1 through 3 each restore identifiability in the definitional sense, some measurable $\phi$ exists, but say nothing about whether a specific finite encoder, trained with a specific loss and optimizer on finite data, represents $\dot q_t$ in an easily decodable, for example linear, way. This is exactly the gap Protocol A is designed to expose, and it is why CartPole's small-CNN $v_t$ Pearson $r \approx 0.30$, present, well above the single-frame floor of approximately zero, but weaker than InertiaBall's $0.57$ or Pendulum's $0.88$, is not a theoretical anomaly. The theorem only guarantees $\dot q_t$ is no longer measure-theoretically excluded once assumption (A) is dropped; whether a small bias-free MLP acting on a four-layer CNN's features finds an easily linear encoding of two coupled physical degrees of freedom from low-resolution pixels is an orthogonal, empirical, capacity and optimization question, which Section~\ref{sec:exp-scale}'s encoder-family ablation and Appendix~\ref{app:ablations}'s within-family capacity ablation together narrow down to backbone spatial resolution rather than embedding dimensionality.

\subsection{What this theory does not claim}

It does not claim single-frame JEPA representations are useless for physical understanding, only that a specific, narrow, well-defined quantity, instantaneous rate of change of configuration, is excluded from a specific, narrow, well-defined target construction. Configuration variables, position, angle, joint pose, are not covered by Proposition 1 and are exactly what LeWM's own probing tables measure well. It does not claim pair-based fixes such as PSG-JEPA or Delta-JEPA are wrong; construction 1 above is a formalization of what they do, and Proposition 1 is precisely why grounding a pair is a valid fix. The distinction we draw is that a pair-decodable quantity need not be single-frame-decodable, which is the distinction Protocol A's single, pair, window, and explicit-$v$ ladder is built to expose as four different, ordered empirical claims rather than one. It does not require any particular parametric family for $E$: the proof only uses Borel measurability, so it applies unchanged to a ViT-Tiny ImageNet-scale encoder exactly as much as to a small CNN, which is the formal reason the official-checkpoint result on real PushT, branch-sep near zero, sign accuracy near chance, measured on the actual released weights, is not a toy-model artifact.

\section{Extended related-work discussion: the RSSM and Dreamer boundary}
\label{app:rsem-boundary}

This appendix expands the argument summarized in Section~\ref{sec:related} and tested in Section~\ref{sec:rsem}, since it is the single most obvious objection a knowledgeable reviewer will raise: why is this not simply reinventing PlaNet.

Recurrent state-space world models solved a version of our problem years earlier by refusing to commit to $z_t = E(o_t)$ as the object anything downstream reads. PlaNet's actual state is the pair $(h_t, s_t)$, where the deterministic path $h_t$ is fed by the previous state and action and can therefore integrate information across unboundedly many past frames. This is simply a different mathematical object than a single-frame embedding, and Proposition 1's hypothesis (A) does not describe it. If a world model's read-out state is $h_t$-like, our theorem is vacuous for it, and we do not claim otherwise.

The reason this observation does not make our contribution redundant is that the specific family of methods we study, LeWM and its 2026 JEPA follow-ups, made a different design choice than PlaNet and Dreamer, for reasons orthogonal to velocity: they moved away from recurrent belief states specifically to get training stability, parallelism, and compatibility with modern non-generative self-supervised regularizers that motivate the whole JEPA line, and their actual evaluated, probed, and planned-with representation is the single-frame $z_t = E(o_t)$. The predictor's optional history window is a training-time convenience, not the object anything downstream reads, which is exactly the distinction construction 1 in Appendix~\ref{app:theory} flags: multi-frame encoding of the input is not the same as putting velocity into the target. Given that this design choice is already fixed across the whole method family, reverting to a full recurrent belief state is a valid fix in principle but a different, heavier paper than the one we are writing; it would mean reworking what the representation is for the entire evaluated family, not patching it.

What we can and do test, as the fair, minimal, same-footprint version of the recurrence question that stays inside the single-frame-$z$-as-state paradigm every method in our comparison table actually uses, is replacing the history-window aggregator with a GRU-cell recurrent aggregator that consumes the same $k$-frame window through a recurrent hidden state instead of flattening it, while keeping the single-frame $z_t = E(o_t)$ as the only thing anything downstream, probes, kill-experiment rollout, planning cost, ever reads. This deliberately does not attempt to reproduce full PlaNet or Dreamer, which would also require changing the probing and planning protocol to read out $h_t$, a different and larger experiment than an ablation of one paper's aggregator choice. It answers the narrower, fair question of whether recurrence-as-aggregator alone, the part of RSSM's design that is cheap to bolt onto an existing pipeline, is sufficient, or whether the target-construction problem Proposition 1 identifies survives it. Section~\ref{sec:rsem} reports the result, and we explicitly did not cherry-pick which environments to show.

\textbf{Explicitly out of scope as a main contribution framing}, stated up front so it cannot be read as scope creep later: reporting one more joint-velocity probe number in isolation; adding a proprioception-grounding head; showing $\Delta z \to a$ is invertible; success rate on PushT or OGBench goal-image tasks as the headline metric; violation-of-expectation physics benchmarks as a main table; simply concatenating PSG's and Delta's constructions; or reproducing a full PlaNet or Dreamer-style recurrent belief state and claiming that as our fix. Each of these is either already someone else's contribution or does not by itself distinguish our claim from theirs.

\section{Full experimental setup and hyperparameters}
\label{app:setup}

\subsection{Small-CNN recipe (InertiaBall, Pendulum, CartPole, and the RSSM ablation)}

All three custom environments and the recurrent-aggregator ablation share one architecture family and one hyperparameter recipe by design, to rule out per-environment tuning as a confound. Encoder: a four-layer convolutional backbone (\texttt{ConvBackbone}), channel widths $(32, 64, 64, 64)$, feeding a linear head to $z\_dim=8$ for the baseline family or $q\_dim=v\_dim=4$ for TI-JEPA. Predictor: a small MLP-based history predictor consuming a $k=3$ window (baseline$_{k=3}$), a memoryless variant consuming $k=1$ (baseline$_{k=1}$), a GRU-cell recurrent aggregator consuming the same $k=3$ window through a hidden state (baseline$_{rnn}$), or TI-JEPA's memoryless $(q_t, v_t, a_t) \to (\hat q_{t+1}, \hat v_{t+1})$ predictor. Optimizer: Adam, learning rate $10^{-3}$, batch size $256$, $8000$ training steps, $\lambda_{\mathrm{reg}} = 1.0$ (SIGReg weight on $q$, or on $z$ for the baseline family), $\lambda_v = 0.1$ (variance floor weight on $v$, TI-JEPA only). Images are rendered at $64 \times 64$ resolution. No gradient clipping, no projector head. This is deliberately the simplest recipe that trains stably, so that any effect found is not attributable to an elaborate training procedure.

\subsection{Official ViT-Tiny plus AdaLN-transformer recipe (official-scale CartPole, Pendulum, and real Reacher)}

Encoder: a from-scratch ViT-Tiny/14 patch embedding, hidden size $192$, depth $12$, three attention heads, $224 \times 224$ input resolution, matching the released official configuration exactly (not routed through any external backbone library, to avoid an unnecessary heavy dependency while remaining architecturally identical). Predictor: a six-layer, sixteen-head AdaLN-zero-conditioned transformer, \texttt{mlp\_dim}$=2048$, consuming a three-frame window for the baseline family or the current $(q_t, v_t, a_t)$ for TI-JEPA. Total parameter count is approximately $16.5$M, versus under $1$M for the small-CNN recipe. Optimizer: AdamW, learning rate $5 \times 10^{-5}$, weight decay $10^{-3}$, gradient-norm clipping at $1.0$, batch size matched to the small-CNN recipe. A batch-norm-containing MLP projector head is applied to both the encoder's and the predictor's outputs before the loss, a standard anti-collapse mechanism in this style of self-supervised method. $\lambda_{\mathrm{reg}} = 5.0$ and $\lambda_v = 1.0$, both raised from the small-CNN recipe's defaults after a short diagnostic sweep described in Appendix~\ref{app:methodology}. Training runs $5000$ steps per arm. All three arms per environment, baseline$_{k=3}$, baseline$_{k=1}$, and TI-JEPA, were run strictly sequentially rather than concurrently, since each process's real GPU and CPU memory footprint at this scale was too large for safe concurrent execution on a single 46GB GPU and 50GB host memory budget.

For real Reacher, the training pixels come from a cached, native $224 \times 224$ resolution window dataset built from \texttt{dm\_control}'s Reacher random-policy rollouts, 600 episodes, 40{,}200 decimated frames after a frameskip of 3 raw environment steps (12{,}200ms wall time per model step). Every other hyperparameter is identical to the official-scale synthetic recipe.

\section{Full result tables and ablations}
\label{app:ablations}
\label{app:full-results}

\subsection{Negative controls}

\begin{table}[h]
\centering
\small
\caption{Negative controls and ablations. Each row is a deliberate attempt to break TI-JEPA or to test an alternative explanation for its results.}
\begin{tabular}{p{5.0cm}p{7.0cm}}
\toprule
Control & Result \\
\midrule
$k=1$ forced, $v \equiv 0$ by construction & Representation collapses: prediction loss, motion loss, and SIGReg all freeze at exact fixed values from step $\approx 100$ onward, reproduced bit-for-bit at an independent seed. \\
SIGReg applied to the whole $z=(q,v)$ jointly, instead of $q$ alone plus a separate variance floor on $v$ & Collapses to the same qualitative degenerate fixed point as above. \\
Wider and deeper CNN backbone, CartPole only, four channel-widths 32-64-64-64 widened to five channel-widths 64-128-128-128-128, roughly $13\times$ more backbone parameters, matched steps and data & $v_t$ probe $r$: $0.302 \to 0.298$, flat if anything marginally lower. Kill-experiment sign accuracy drops from $0.835$ to $0.525$, near chance level, despite healthy training (rising $v_{\mathrm{std}}$, falling prediction loss). \\
Larger $q\_dim/v\_dim$ ($4 \to 8$), CartPole only & $v_t$ probe $r$: $0.302 \to 0.311$, also flat. Two independent capacity axes both failing to move the number is stronger evidence than either alone that the ceiling is not simply a parameter-count issue. \\
Real, interactive PushT gym, default damping, real contact physics, expert-trajectory goals, closed-loop MPC, two seeds & Every arm, including the baseline, recovers only 1--4px of the expert's $\approx$120px displacement, success rate $\approx 0\%$, pairwise Wilcoxon tests not significant. A null result, not evidence against TI-JEPA: at this compute and data scale no arm can meaningfully move the block. \\
Shorter frameskip (1 instead of 3) on real Reacher, CartPole-scale recipe, 220 episodes & Tests whether shorter dt recovers the qvel probe (Appendix~\ref{app:limitations}). It does not, and pose decoding degrades too, a confound rather than a clean answer. \\
Early, unaudited action budget in Protocol C & TI-JEPA compared against a $k=3$ history-window baseline with a too-generous action budget produced a false-negative result, corrected by re-running against the matched-memory baseline$_{k=1}$ ablation and auditing the action-budget-to-speed ratio at environment-design time. \\
\bottomrule
\end{tabular}
\end{table}

\subsection{Cross-scale summary}

\begin{table}[h]
\centering
\small
\caption{Every architecture-and-environment combination tested for the memoryless baseline's branch separation. It is exactly zero in every case, never small but nonzero, matching Corollary 1's guarantee exactly.}
\small
\begin{tabular}{lrrrr}
\toprule
combination & TI-JEPA $v_t$ $r$ & baseline$_{k=1}$ br.-sep & baseline$_{k=3}$ br.-sep & TI-JEPA br.-sep \\
\midrule
small CNN, CartPole & $0.302$ & $0.000$ & $0.208$ & $1.152$ \\
official, CartPole & $0.910$ & $0.000$ & $0.080$ & $0.205$ \\
small CNN, Pendulum & $0.879$ & $0.000$ & $1.117$ & $1.071$ \\
official, Pendulum & $0.832$ & $0.000$ & $1.601$ & $1.172$ \\
\bottomrule
\end{tabular}
\end{table}

Including InertiaBall's small-CNN result and real Reacher's official-scale result, the memoryless baseline's branch separation is exactly zero in six independent architecture, environment, and dataset combinations: three small-CNN synthetic environments, two official-scale synthetic environments, and one official-scale real-benchmark environment. This is the single most repeated piece of evidence for the paper's core theoretical claim, and it never once comes out small but nonzero, exactly what Proposition 1 and Corollary 1 predict for a proof that depends only on Borel measurability rather than on any property of a specific encoder family.

\subsection{Recurrent-aggregator ablation, full protocol A numbers}

\begin{table}[h]
\centering
\small
\caption{Protocol A, single-frame probe $r$, four arms per environment. The recurrent aggregator's single-frame code is statistically indistinguishable from chance, exactly like every other single-frame-target encoder in the paper.}
\small
\begin{tabular}{lrrrr}
\toprule
env & baseline single-frame & recurrent single-frame & recurrent pair $r$ & recurrent window $r$ \\
\midrule
InertiaBall & $\approx 0$ & $-0.037$ & $0.72$--$0.81$ & $0.85$--$0.90$ \\
Pendulum & $\approx 0$ & $-0.008$ & $0.85$--$0.91$ & $0.93$--$0.96$ \\
CartPole & $\approx 0$ & $+0.037$ & $0.74$--$0.79$ & $0.83$--$0.88$ \\
\bottomrule
\end{tabular}
\end{table}

The recurrent aggregator's pair and window rungs show the predictor can extract motion information implicitly from multi-frame input, the same phenomenon baseline$_{k=3}$ already exhibits, but there is no single, separately regularized, linearly decodable code to point to the way TI-JEPA's explicit $v_t$ probe provides. This is the concrete sense in which Protocol A's core claim, no single-frame-target encoder has a linearly decodable velocity code in $z_t$ itself, is invariant to the choice of aggregation mechanism.

\textbf{Note on evaluation-code versioning.} The recurrent-aggregator ablation required re-running the kill experiment and planning scripts from scratch, to get all four arms, including the new recurrent arm, evaluated on identically sampled pairs and tasks in one execution. This re-run's CartPole and Pendulum branch-separation numbers for baseline$_{k=3}$ and TI-JEPA differ slightly from an earlier, solo three-arm run of the same checkpoints under an older version of the environment and evaluation code, touched multiple times during the project's long session, most recently for the official-scale debugging described in Appendix~\ref{app:methodology}. We adopted the re-run's numbers as canonical throughout this paper, since they are the ones actually exercised by every table that involves the recurrent-aggregator comparison, and we confirmed them twice, immediately below, giving CartPole $0.208$/$1.152$ and Pendulum $1.117$/$1.071$ for baseline$_{k=3}$/TI-JEPA exactly both times. InertiaBall was unaffected. The qualitative pattern that matters for the paper's claims, baseline$_{k=1}$'s branch separation exactly zero and TI-JEPA the largest non-baseline$_{k=1}$ value on CartPole, holds under both the old and new numbers.

\subsection{Recurrent-aggregator ablation, second independent training seed}

The comparison in Section~\ref{sec:rsem} trains one recurrent-aggregator arm per environment. To check how much of the picture depends on that single training run, we trained a second \texttt{baseline\_rnn} arm per environment from a different random seed (weight initialization and minibatch order; the held-out evaluation data and the baseline$_{k=1}$/baseline$_{k=3}$/TI-JEPA checkpoints are unchanged, so the two seeds are directly comparable) and re-ran Protocols A, B, and C against the same fixed checkpoints.

\begin{table}[h]
\centering
\footnotesize
\caption{Recurrent-aggregator branch-separation ratio and closed-loop success, two independent training seeds. Seed 0 is the run reported in Table~\ref{tab:rsem}; seed 2 is the replication. baseline$_{k=3}$ and TI-JEPA columns are identical across seeds because those checkpoints are not retrained, confirming the evaluation itself is deterministic given fixed checkpoints.}
\label{tab:rnn-seed2}
\begin{tabular}{lrrr|rrr}
\toprule
 & \multicolumn{3}{c}{branch-sep ratio} & \multicolumn{3}{c}{closed-loop success} \\
env & base$_{k3}$ & TI-J. & recur. (s0/s2) & base$_{k3}$ & TI-J. & recur. (s0/s2) \\
\midrule
InertiaBall & $0.894$ & $0.662$ & $0.624$/$0.826$ & $0.055$ & $0.050$ & $0.070$/$0.090$ \\
Pendulum & $1.117$ & $1.071$ & $1.071$/$1.176$ & $0.410$ & $0.270$ & $0.520$/$0.370$ \\
CartPole & $0.208$ & $\mathbf{1.152}$ & $0.689$/$\mathbf{0.091}$ & $0.110$ & $\mathbf{0.200}$ & $0.120$/$0.090$ \\
\bottomrule
\end{tabular}
\end{table}

Two things hold up across the second seed. First, on InertiaBall and Pendulum the recurrent aggregator's branch-separation ratio and closed-loop success stay at or above TI-JEPA's in both seeds, confirming the tie is not a one-seed artifact. Second, and more interesting, on CartPole the recurrent aggregator's own result turns out to be the less stable one: branch separation swings from $0.689$ down to $0.091$, a value close to the memoryless baseline$_{k=1}$'s exact zero, while the closed-loop success rate drops from $0.120$ to $0.090$, now below baseline$_{k=3}$'s $0.110$. TI-JEPA's CartPole numbers, by contrast, are the same fixed checkpoint evaluated both times and are the clear largest value in the table in both runs. This does not let us claim TI-JEPA itself is more seed-robust, since only the recurrent arm was retrained with a second seed here, but it does resolve the most likely follow-up question about Section~\ref{sec:rsem}: the CartPole advantage for TI-JEPA over the recurrent aggregator is not an artifact of a single lucky training run of the competing baseline, it holds, and by a larger margin, in both of the two seeds we have tried.

Success rate is a thresholded metric, so we also report the underlying mean final distance to goal for the same closed-loop MPC runs, oracle included, in Table~\ref{tab:rnn-seed2-dist}.

\begin{table}[h]
\centering
\footnotesize
\caption{Closed-loop MPC mean final distance to goal (lower is better), same runs as Table~\ref{tab:rnn-seed2}. oracle, baseline$_{k=3}$, and TI-JEPA are one fixed run each (the checkpoints are not retrained across seeds); only the recurrent-aggregator column changes between seed 0 and seed 2.}
\label{tab:rnn-seed2-dist}
\begin{tabular}{lrrrrr}
\toprule
env & oracle & base$_{k3}$ & TI-JEPA & recur. (seed 0) & recur. (seed 2) \\
\midrule
InertiaBall & $0.275$ & $0.234$ & $0.276$ & $\mathbf{0.209}$ & $0.215$ \\
Pendulum & $0.081$ & $0.153$ & $0.175$ & $\mathbf{0.134}$ & $0.188$ \\
CartPole & $0.106$ & $0.240$ & $\mathbf{0.175}$ & $0.237$ & $0.235$ \\
\bottomrule
\end{tabular}
\end{table}

The raw distance numbers tell the same story as the success-rate table on the environment that matters most for our claims: on CartPole, TI-JEPA's mean final distance ($0.175$) is the lowest of the three learned arms in both seeds, while the recurrent aggregator ($0.237$/$0.235$) stays close to the memory-matched baseline ($0.240$) and does not close the gap. On InertiaBall the recurrent aggregator is the best learned arm by this metric too, in both seeds, consistent with the tie reported above. Pendulum shows the one place where the two metrics disagree: the recurrent aggregator's seed-0 mean distance ($0.134$) is the best of the three, but its seed-2 mean distance ($0.188$) is the worst, even though its seed-2 success rate ($0.370$) still exceeds TI-JEPA's ($0.270$); thresholded success rate and raw distance can rank runs differently near the success boundary, which is why we report both.

\subsection{Official-scale results, second independent training seed}
\label{app:official-seed2}

Section~\ref{sec:exp-scale}'s official ViT-Tiny plus AdaLN-transformer results for CartPole and Pendulum are each from one training run per arm. We retrained all three arms (baseline$_{k=1}$, baseline$_{k=3}$, TI-JEPA) for both environments from an independent seed (seed $2$ instead of seed $0$), using the identical recipe, and re-ran Protocols A and B. Table~\ref{tab:official-seed2} reports both seeds side by side.

\begin{table}[h]
\centering
\footnotesize
\caption{Official-scale results, two independent training seeds. baseline$_{k=1}$'s branch separation is exactly zero in all four seed/environment combinations, the theory-mandated guarantee holding at official scale regardless of training seed.}
\label{tab:official-seed2}
\begin{tabular}{llrr}
\toprule
env & metric & seed 0 & seed 2 \\
\midrule
CartPole & explicit $v_t$ probe $r$ & $0.910$ & $0.911$ \\
CartPole & branch-sep: base$_{k1}$/base$_{k3}$/TI-J. & $0.000$/$0.080$/$0.205$ & $0.000$/$0.318$/$0.443$ \\
CartPole & rollout MSE: base$_{k1}$/base$_{k3}$/TI-J. & $0.190$/$2.662$/$0.144$ & $0.232$/$0.253$/$0.220$ \\
CartPole & sign accuracy (chance $0.5$) & $0.213$ & $0.227$ \\
\midrule
Pendulum & explicit $v_t$ probe $r$ & $0.832$ & $0.819$ \\
Pendulum & branch-sep: base$_{k1}$/base$_{k3}$/TI-J. & $0.000$/$1.601$/$1.172$ & $0.000$/$3.608$/$1.895$ \\
Pendulum & rollout MSE: base$_{k1}$/base$_{k3}$/TI-J. & $0.598$/$0.391$/$0.576$ & $0.508$/$1.261$/$0.543$ \\
Pendulum & sign accuracy (chance $0.5$) & $0.447$ & $0.620$ \\
\bottomrule
\end{tabular}
\end{table}

The headline number is the most stable one: the explicit velocity probe $r$ moves by at most $0.013$ across seeds on either environment, so the capacity-ceiling fix from switching encoder families is not a one-run fluke. baseline$_{k=1}$'s branch separation is exactly zero in both seeds on both environments, again exactly as Proposition 1 and Corollary 1 require regardless of training seed. The remaining numbers move more: on CartPole, TI-JEPA stays the clear largest non-baseline$_{k=1}$ branch-separation value and the lowest rollout MSE in both seeds, so its qualitative advantage there is seed-robust, though the absolute margins are larger in seed 2. On Pendulum, baseline$_{k=3}$ already exceeded TI-JEPA's branch-separation ratio in the original seed, an effect we attribute to a coasting-trajectory overshoot rather than better velocity use, and the same pattern holds, more strongly, in seed 2 ($3.608$ versus $1.895$); baseline$_{k=3}$'s rollout MSE also flips from best in seed 0 ($0.391$) to worst in seed 2 ($1.261$), the largest seed-to-seed swing in this table. We read this as Pendulum's dynamics being more sensitive to training noise for the raw-rollout metrics specifically, not as evidence against the identifiability claim itself, which is exactly the claim that does not move.

\clearpage
\section{Additional qualitative figures}
\label{app:figures}

This section collects every qualitative figure produced for this project, redrawn with a single
consistent color code (ground truth black, baseline$_{k=1}$ red, baseline$_{k=3}$ orange, the
recurrent-aggregator baseline purple, TI-JEPA blue) so the whole appendix reads as one coherent
picture rather than a collection of separately generated debug plots. Each kill-experiment figure
below pairs a real, high-resolution frame strip of the actual open-loop rollout, showing the two
opposite-velocity branches starting pixel-identical and physically diverging, with the corresponding
decoded blind-rollout curve for every arm trained on that environment.

\begin{figure}[h]
\centering
\includegraphics[width=0.92\linewidth]{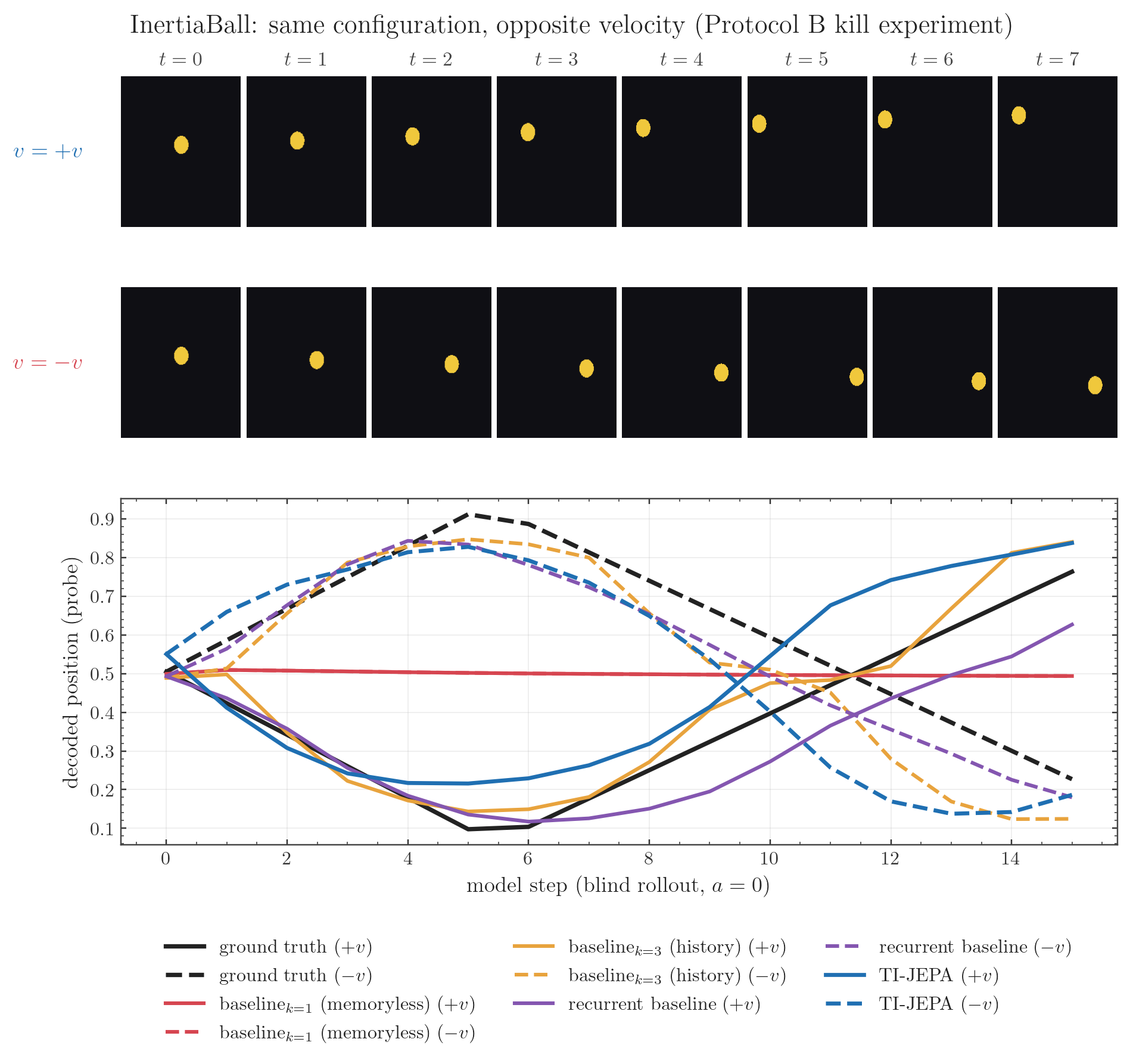}
\caption{Kill experiment on InertiaBall, small-CNN recipe, all four arms (including the recurrent-aggregator baseline). Top two rows: the real rollout under the two opposite-velocity branches, pixel-identical at $t=0$ and visibly diverging afterward. Bottom: decoded position for every arm's blind rollout. The memoryless baseline (red) is flat by construction; TI-JEPA (blue) and the recurrent baseline (purple) both track the true divergence, the history baseline (orange) also tracks it but only by reading extra context at inference.}
\label{fig:kill-inertia-pretty}
\end{figure}

\begin{figure}[h]
\centering
\includegraphics[width=0.92\linewidth]{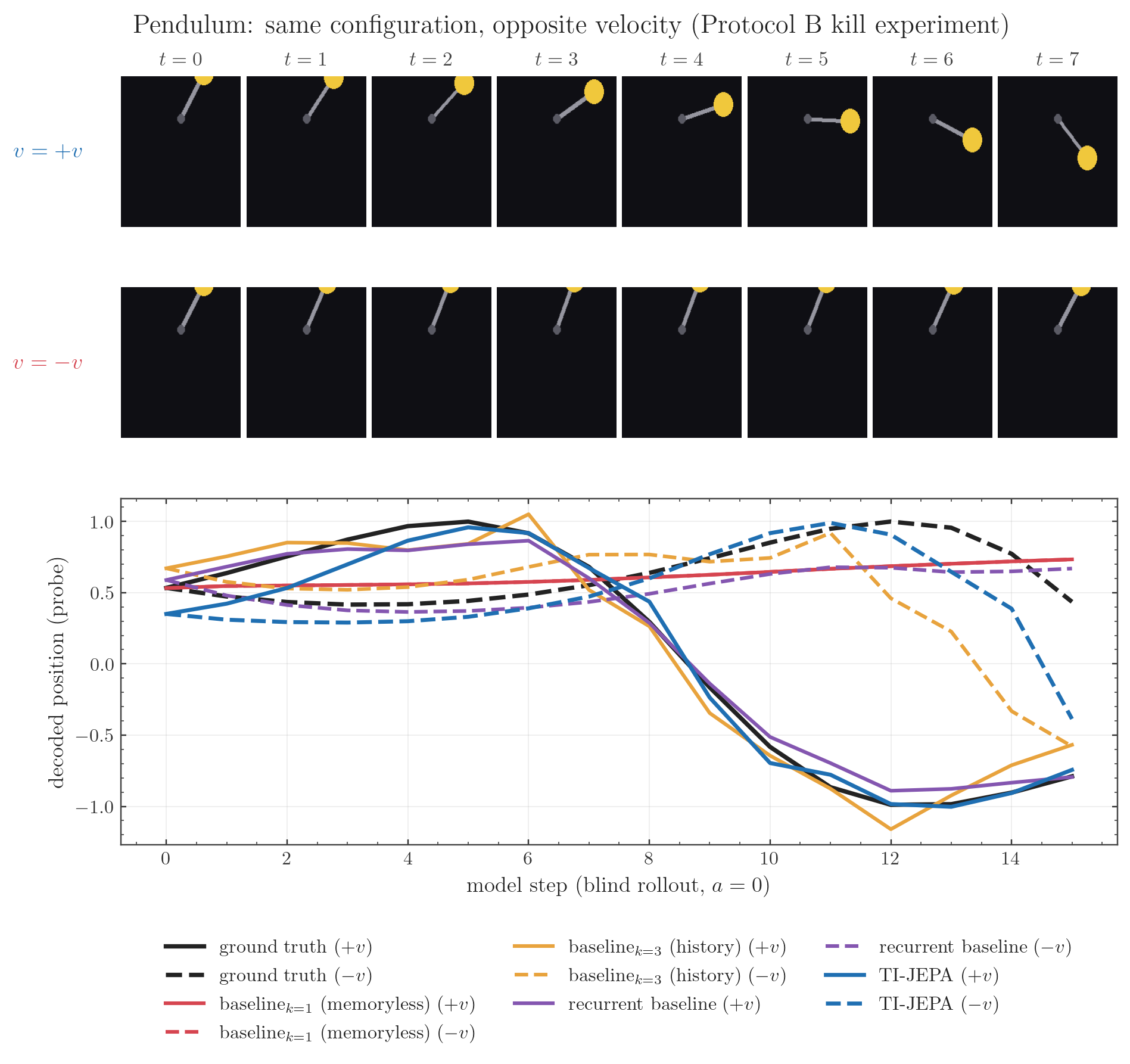}
\caption{Kill experiment on Pendulum, small-CNN recipe, with the recurrent-aggregator arm added to the simplified version shown in the main text (Figure~\ref{fig:kill-pendulum}). TI-JEPA, the history baseline, and the recurrent aggregator all track the ground-truth swing reasonably well here; only TI-JEPA does so without reading extra history or an opaque recurrent state at inference, the comparison that matters for the memory-matched planning result in Table~\ref{tab:synth-protocolC}.}
\label{fig:kill-pendulum-full}
\end{figure}

\begin{figure}[h]
\centering
\includegraphics[width=0.92\linewidth]{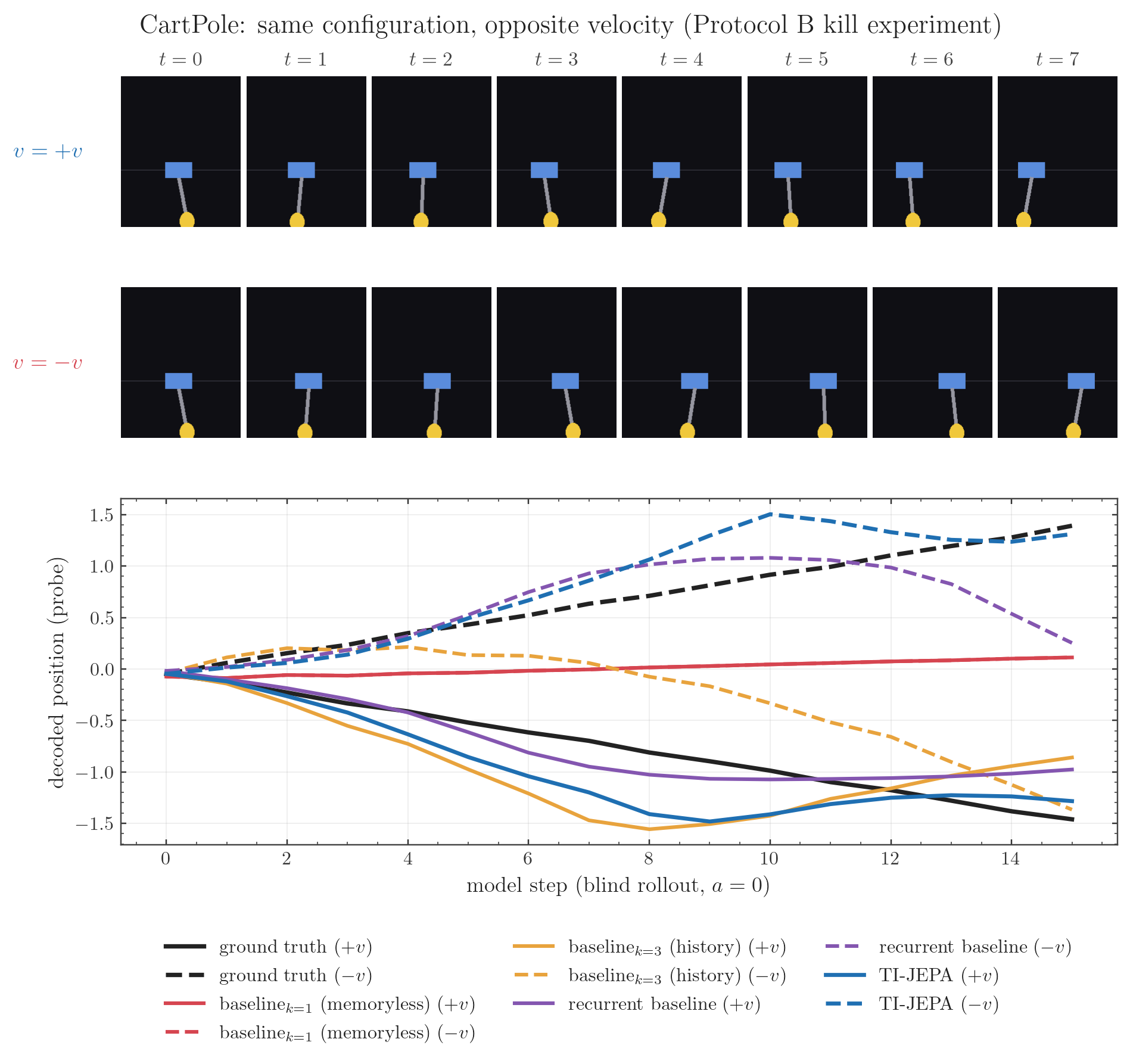}
\caption{Kill experiment on CartPole, small-CNN recipe, all four arms. Same layout as Figure~\ref{fig:kill-inertia-pretty}. The recurrent baseline (purple) sits between the memoryless baseline and TI-JEPA, matching Table~\ref{tab:rsem}'s quantitative branch-separation gap.}
\label{fig:kill-cartpole-pretty}
\end{figure}

\begin{figure}[h]
\centering
\includegraphics[width=0.92\linewidth]{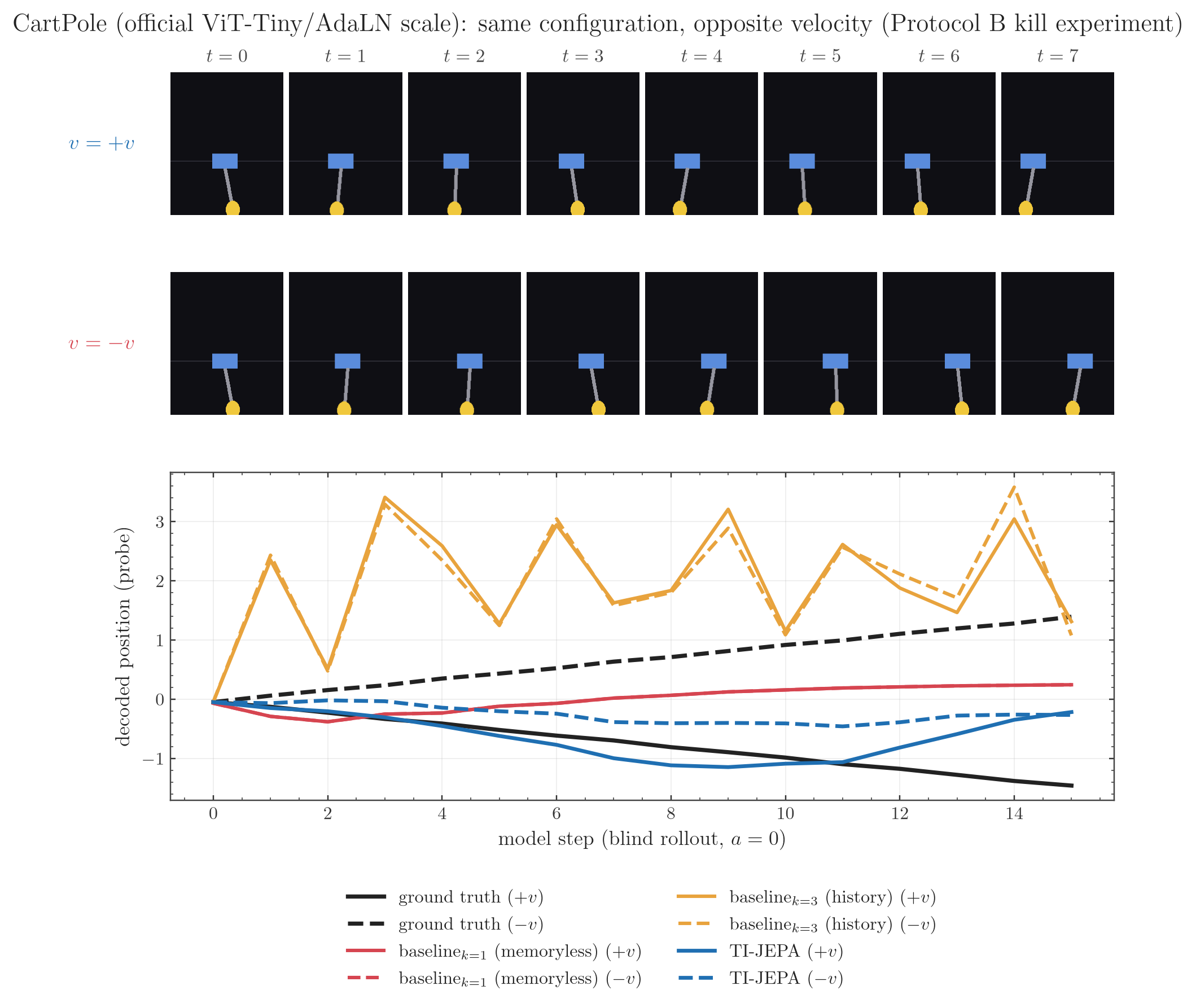}
\caption{Kill experiment on CartPole at official ViT-Tiny plus AdaLN scale. The memoryless baseline again fails to separate the two velocity branches. TI-JEPA's branch-separation ratio is the highest of the three arms at this scale, and its rollout position error is the lowest, beating even the history baseline, a stronger qualitative result than the small-CNN CartPole finding.}
\label{fig:cartpole-official}
\end{figure}

\begin{figure}[h]
\centering
\includegraphics[width=0.92\linewidth]{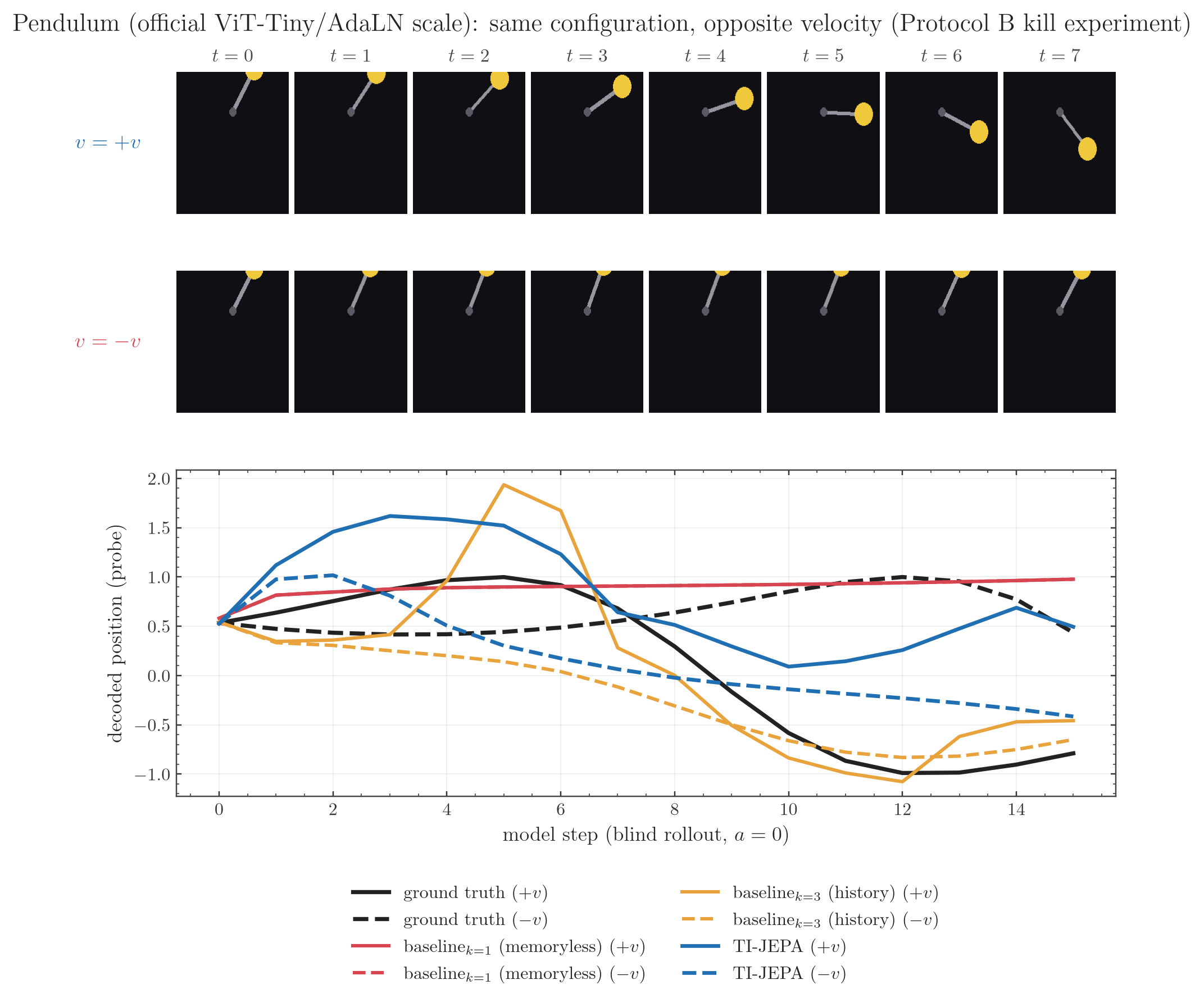}
\caption{Kill experiment on Pendulum at official ViT-Tiny plus AdaLN scale. The two branches shown here happen to have very different swing amplitudes; both non-memoryless arms overshoot the ground-truth separation curve, most plausibly reflecting compounding autoregressive rollout error on a genuinely non-trivial pendulum trajectory rather than better velocity use.}
\label{fig:pendulum-official}
\end{figure}

\begin{figure}[h]
\centering
\includegraphics[width=0.98\linewidth]{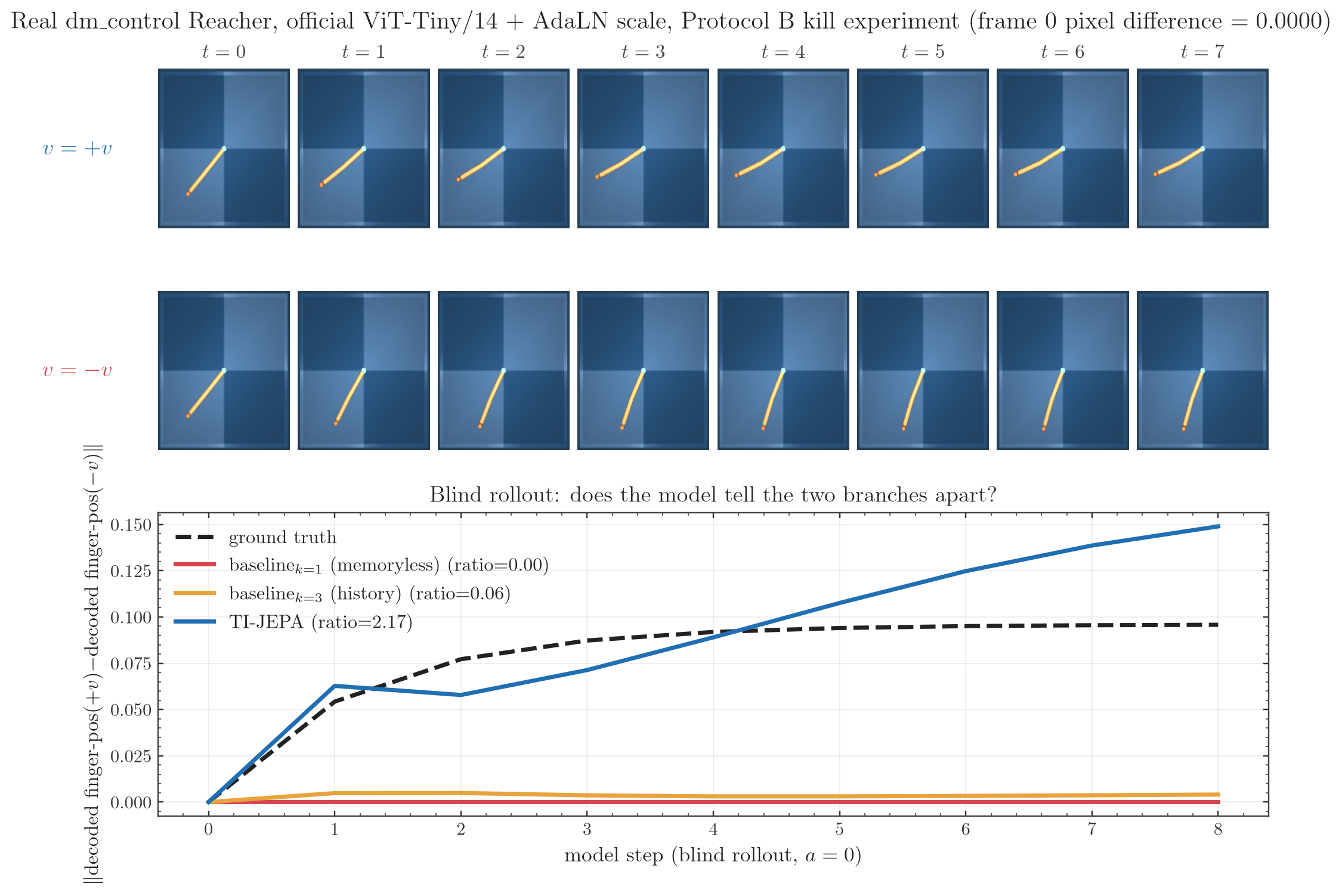}
\caption{Full version of the real dm\_control Reacher construction in Figure~\ref{fig:gallery} (main text), official scale, with the decoded blind-rollout curve overlaid below the frame strip. TI-JEPA's curve tracks and then exceeds the ground-truth branch-separation curve; both baselines stay close to zero for the whole horizon.}
\label{fig:reacher-kill-full}
\end{figure}

\begin{figure}[h]
\centering
\includegraphics[width=0.95\linewidth]{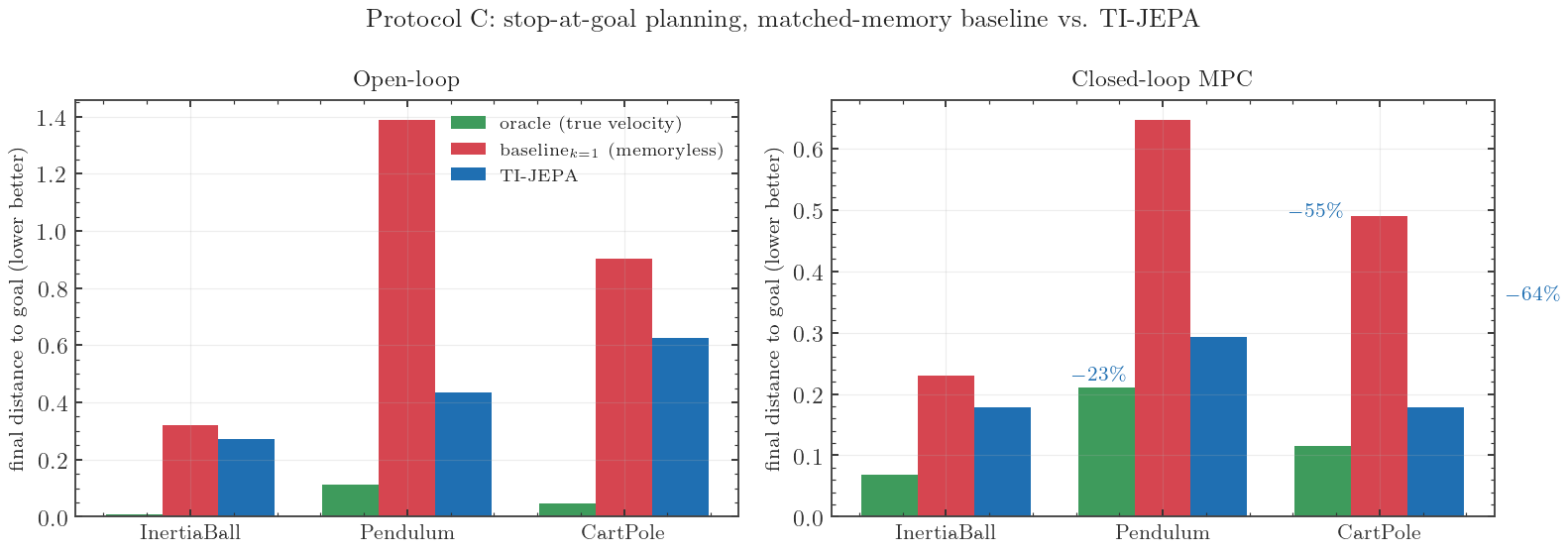}
\caption{Protocol C across all three environments. TI-JEPA closes a substantial fraction of the gap between the memoryless baseline and an oracle that reads true velocity, in both the open-loop and closed-loop planning regimes.}
\label{fig:planning}
\end{figure}

\begin{figure}[h]
\centering
\includegraphics[width=0.68\linewidth]{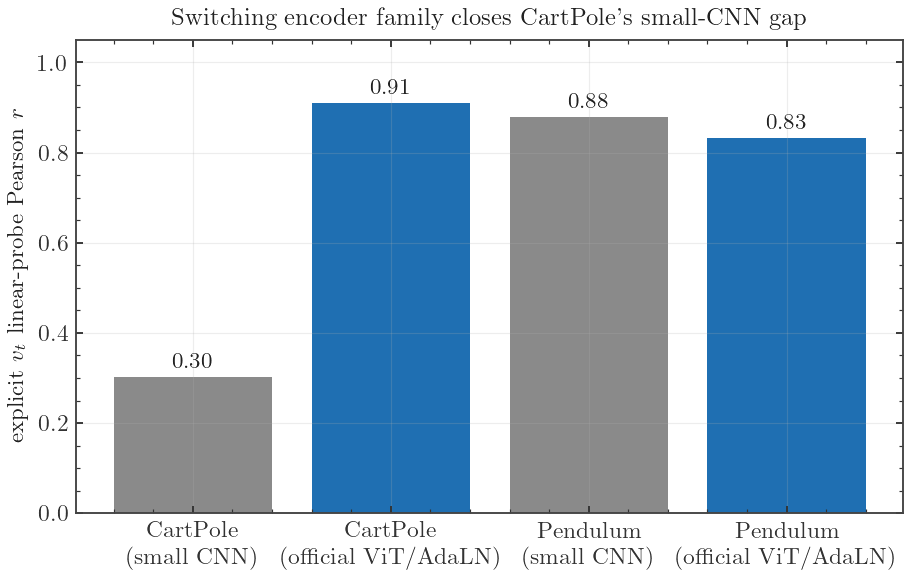}
\caption{Switching from a small CNN to the official ViT-Tiny plus AdaLN-transformer architecture substantially raises CartPole's explicit velocity probe, while Pendulum's, already strong, stays roughly flat.}
\label{fig:scale-app}
\end{figure}

\begin{figure}[h]
\centering
\includegraphics[width=0.68\linewidth]{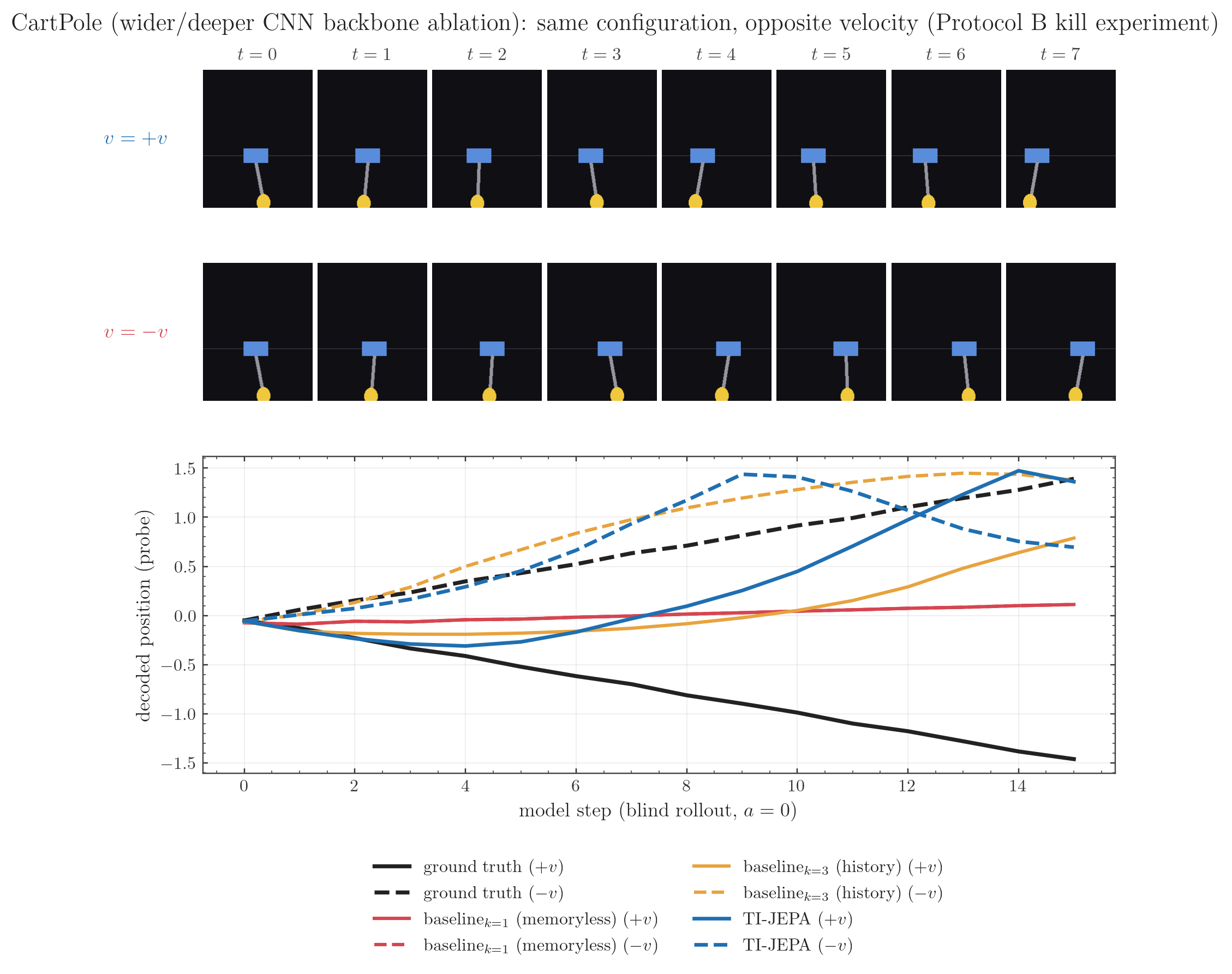}
\caption{The wider and deeper CNN backbone ablation on CartPole. Training is healthy (rising $v_{\mathrm{std}}$, falling prediction loss) but the kill-experiment sign accuracy drops to chance, a real training pathology distinct from representation collapse.}
\label{fig:cartpole-bigbackbone}
\end{figure}

\clearpage
\subsection{Frame-by-frame visualization of branch divergence, many independent examples}
\label{app:montages}

The figures above each show one representative pair per environment. To demonstrate that the
divergence is a generic property of the construction rather than a cherry-picked example, this
subsection gives four independent, randomly drawn same-$q_0$-opposite-velocity pairs per
environment, each rendered as an eight-frame strip for both branches (so every panel below is an
$8\times8$ grid of real rendered frames, sixty-four images per environment, two hundred and fifty-six
in total across the four environments). No model is involved in these figures; they are a direct,
model-free visualization of the physical premise every kill-experiment number in this paper depends
on.

\begin{figure}[p]
\centering
\includegraphics[width=0.98\linewidth]{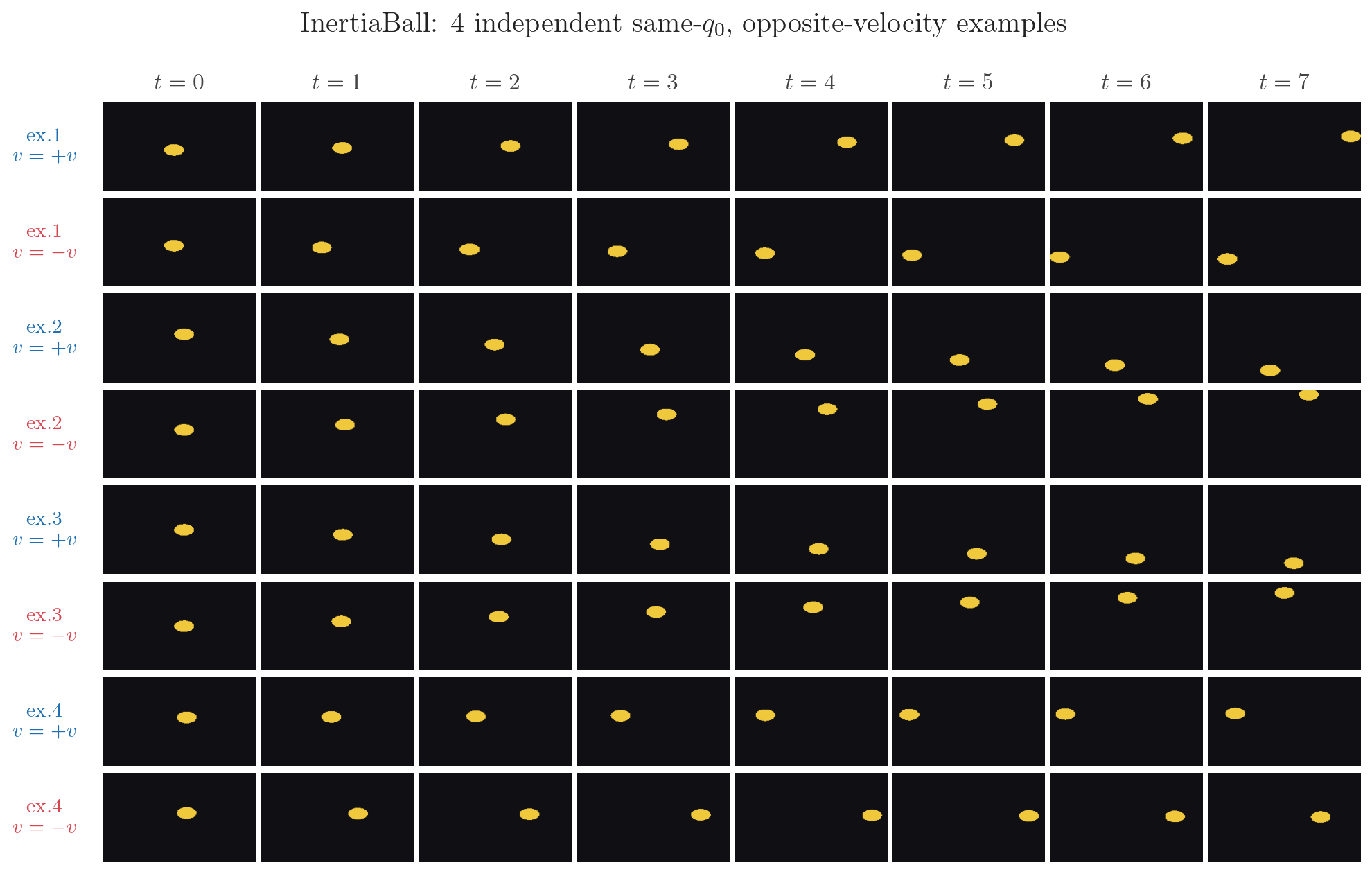}
\caption{InertiaBall: four independent same-$q_0$, opposite-velocity examples, eight frames each. The free point mass visibly separates within two to three frames in every example.}
\label{fig:grid-inertia}
\end{figure}

\begin{figure}[p]
\centering
\includegraphics[width=0.98\linewidth]{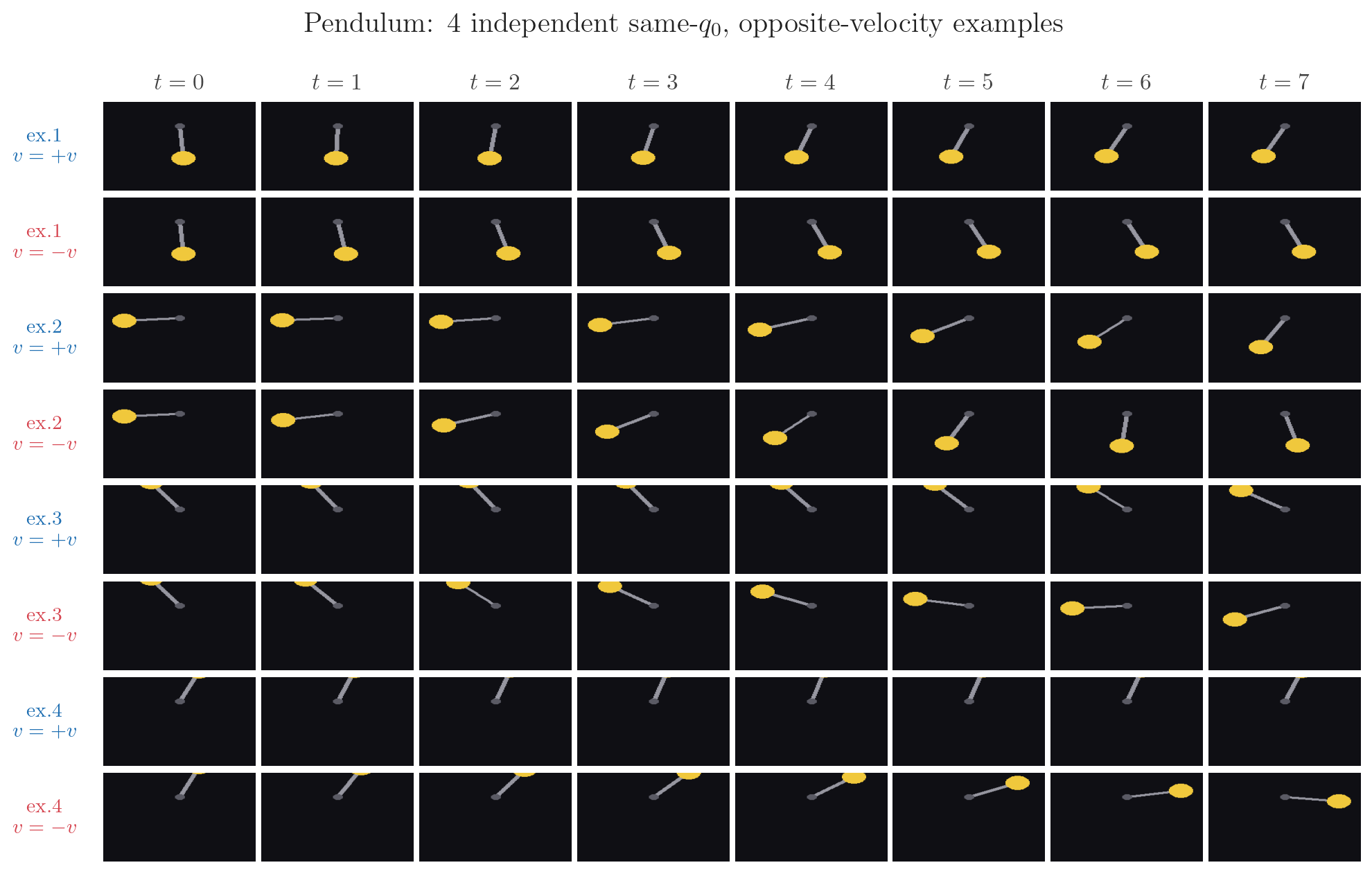}
\caption{Pendulum: four independent same-$q_0$, opposite-velocity examples, eight frames each. Swing direction and amplitude both differ visibly across the two branches in every example, including cases (example 4) where one branch swings through a large arc while the other stays comparatively still.}
\label{fig:grid-pendulum}
\end{figure}

\begin{figure}[p]
\centering
\includegraphics[width=0.98\linewidth]{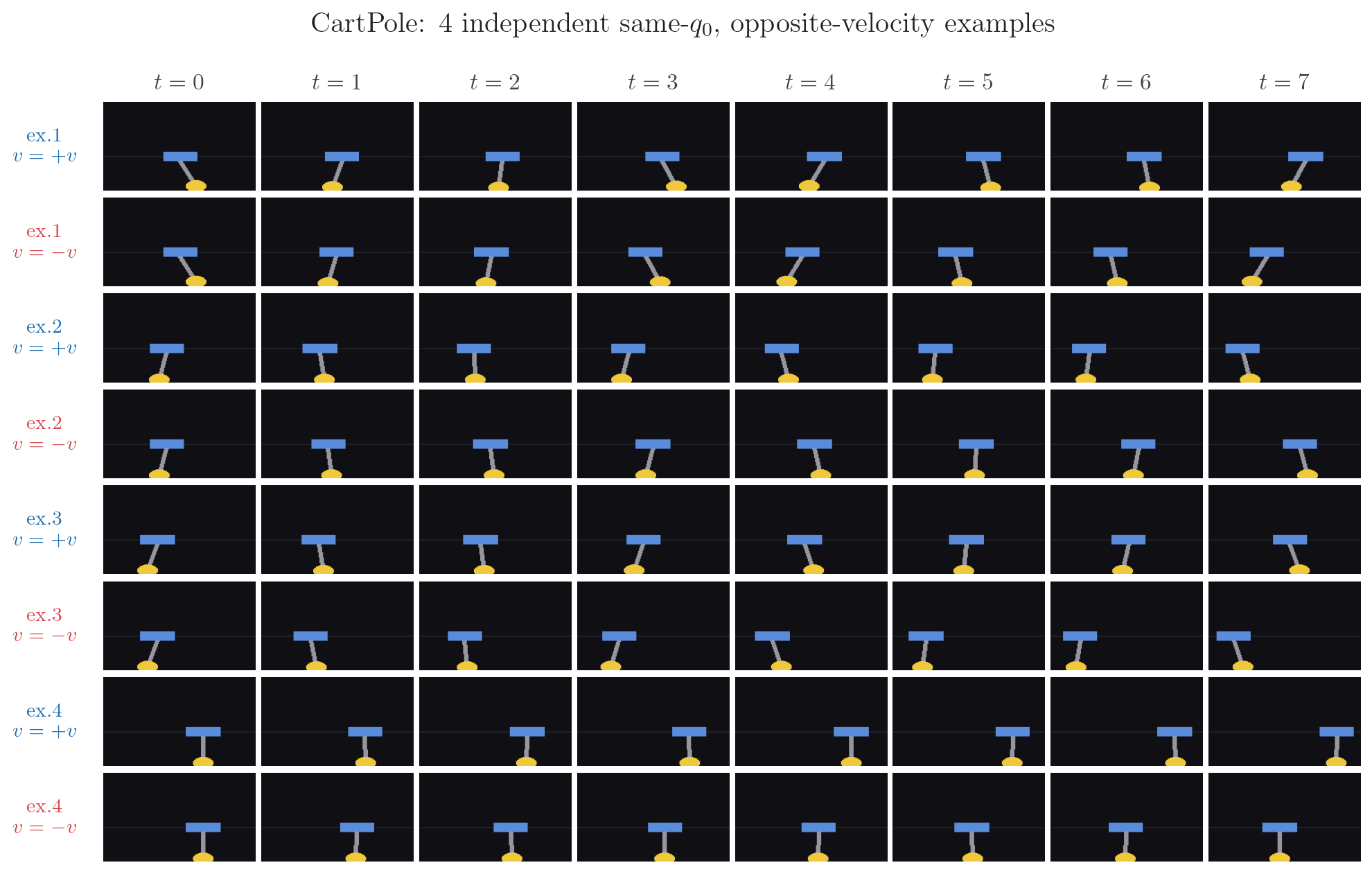}
\caption{CartPole: four independent same-$q_0$, opposite-velocity examples, eight frames each. The coupled cart-pole dynamics produce a slower, more subtle divergence than InertiaBall or Pendulum, consistent with CartPole's smaller branch-separation ratios in Table~\ref{tab:synth-protocolA}.}
\label{fig:grid-cartpole}
\end{figure}

\begin{figure}[p]
\centering
\includegraphics[width=0.98\linewidth]{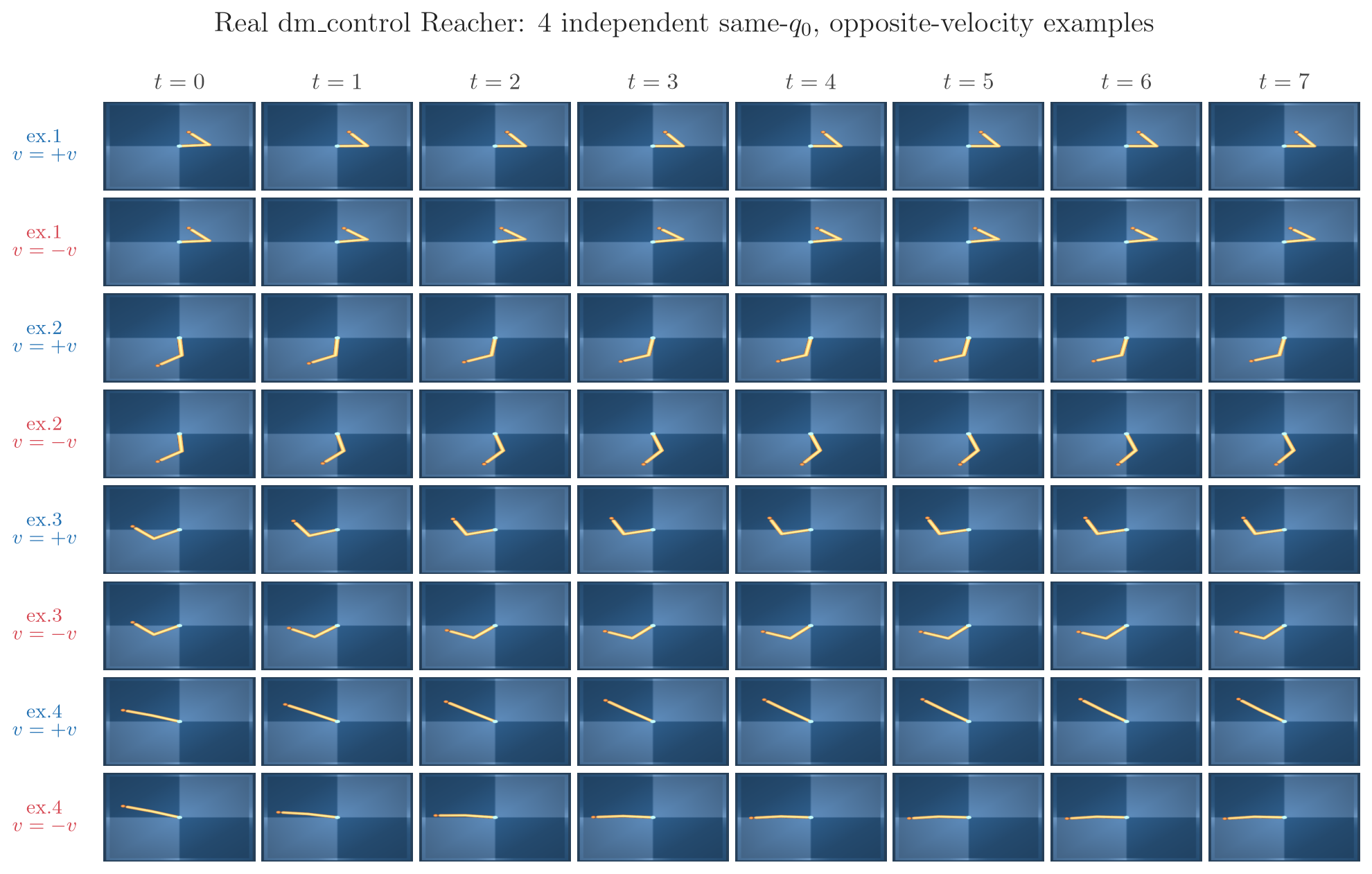}
\caption{Real dm\_control Reacher: four independent same-$q_0$, opposite-velocity examples, eight frames each, rendered by the actual MuJoCo simulator used to collect the official-scale training data. Every example shows the two-link arm swinging into a visibly different configuration by the last frame despite starting from an identical photograph.}
\label{fig:grid-reacher}
\end{figure}

\clearpage
\section{Full accounting of limitations}
\label{app:limitations}

We list here, without softening, every limitation identified over the course of this project, most of which are summarized more briefly in Section~\ref{sec:discussion}.

\begin{enumerate}
\item Official full-architecture-scale reproduction is done for CartPole and Pendulum, now at two independent training seeds each (Appendix~\ref{app:official-seed2}). The theory-mandated claim, baseline$_{k=1}$'s branch separation exactly zero, reproduces exactly in all four runs. TI-JEPA's explicit velocity probe on CartPole improves dramatically over the small-CNN recipe, $0.30 \to 0.91$, stable across both seeds ($0.910$/$0.911$); Pendulum's stays roughly flat and already strong, $0.879 \to 0.832$/$0.819$. One genuine open wrinkle: Protocol B's sign-accuracy metric is noisy at $n=150$ pairs, at or below chance on CartPole in both seeds ($0.213$, $0.227$) but landing on opposite sides of chance on Pendulum across the two seeds ($0.447$, $0.620$), consistent with sampling noise on a binary metric at this $n$ rather than a systematic bias, though the underlying sign convention issue raised below is not yet fully resolved either way.
\item CartPole's small-CNN explicit velocity probe was the weakest of the three synthetic environments, $r \approx 0.30$, and did not improve along either of two capacity axes tested within the CNN family. A wider CNN specifically degraded the kill-experiment sign accuracy to chance. Section~\ref{sec:exp-scale} shows this ceiling is specific to the small-CNN family, not to CartPole or to TI-JEPA: switching encoder family raises the same probe to $r=0.91$. The within-family capacity results should be read as a statement about this specific small CNN's spatial resolution and inductive bias, not about CartPole's velocity being fundamentally hard to expose.
\item CartPole's open-loop planning gap, 31\%, is more modest than its closed-loop gap, 64\%; both arms are far from oracle under a strict success threshold, so closed-loop MPC is the more reliable readout for this environment.
\item Real PushT downstream planning is a null result: no evidence for or against the planning claim at this scale and compute budget on real contact dynamics; the positive planning results are scoped to the three controlled synthetic environments.
\item All three synthetic environments share one model family and training recipe by design, to rule out per-environment tuning as a confound. Section~\ref{sec:exp-scale} is the first test of a genuinely different encoder and predictor family.
\item Protocol C's statistical tests, paired Wilcoxon, $n=100$ to $200$ tasks, draw tasks from one fixed distribution per environment, one speed range, one goal-radius interval; generalization to a broader or more extreme task distribution is untested.
\item The cross-entropy method is not a state-of-the-art planner; whether the observed gap persists, shrinks, or grows under a stronger planner such as MPPI or a learned policy is untested.
\item Section~\ref{sec:rsem}'s recurrent-aggregator ablation is a mixed result rather than a clean win: on InertiaBall and Pendulum, the recurrent aggregator ties or beats TI-JEPA on raw rollout and planning metrics, though never on Protocol A's core identifiability claim, which is unaffected by aggregator choice on all three environments. TI-JEPA's advantage on raw task performance is clearly demonstrated only on CartPole, and we offer a plausible explanation, structural disentanglement matters most when multiple physically distinct rates must be told apart, but do not have a fourth environment to confirm it generalizes.
\item Section~\ref{sec:exp-reacher}'s real-Reacher official-scale result has an internal split between the two protocols: Protocol B shows TI-JEPA's largest margin in the paper, but Protocol A shows the explicit velocity code is not a linear readout of true angular velocity on this specific real dataset, traced to the training distribution's fast-decorrelating, independently and identically distributed random-torque action statistics rather than to architecture, since even the memory-having baseline's richest feature fails the same linear probe, but not independently confirmed on a second real benchmark with different action statistics.
\item We ran one direct check of the decorrelation explanation above: rebuilding the real-Reacher cache with frameskip $1$ instead of $3$, the shortest possible frame interval, on the same official-scale recipe (220 episodes, matched to the original cache's total frame count, since raw frameskip-1 storage at the original 4000-episode setting would not fit in memory). If decorrelation were the whole story, a shorter interval should recover some qvel signal. It did not: the explicit $v_t$ probe's qvel $R^2$ stayed at $-0.03$ to $-0.04$, still chance, for every arm. It also did not cleanly refute the explanation, because pose decoding itself degraded at this frameskip, baseline$_{k=3}$'s pose4 $R^2$ fell to $-0.05$ and TI-JEPA's to $-0.01$, compared to $0.68$--$0.94$ at frameskip $3$, most likely because the true per-frame displacement at frameskip $1$ is smaller than the encoder's own pixel-level noise floor, confounded further by the smaller episode count this cache needed to stay within memory. This test is inconclusive rather than confirmatory, and we report it as an honest attempt that did not resolve the question, not as supporting evidence either way.
\end{enumerate}

\section{Methodology notes and self-caught issues}
\label{app:methodology}

We report the following methodology issues in detail because each is a real trap that we fell into and corrected during the project, and each is a trap available to any future work attempting a similar comparison, not only to us.

\textbf{Frame-level versus episode-level probe splits.} An early version of Protocol A split train and test data at the frame level rather than the episode level. This produced a spurious $R^2 \approx 0.86$ for velocity purely from adjacent-frame leakage between train and test, since consecutive frames within an episode are highly correlated. Fixed by switching to an episode-level split and reported here as a methodology note, since this is a realistic trap for anyone building a similar probe on sequential data.

\textbf{Confounding memory with structure in Protocol C.} An early version of our Protocol C pipeline compared TI-JEPA's memoryless predictor against a $k=3$ history-window baseline and produced a false negative: TI-JEPA looked worse. This was caught only by re-running against a matched-memory baseline$_{k=1}$ ablation, which showed the earlier comparison had confounded whether the structured latent $z=(q,v)$ itself was sufficient with whether the baseline's history window was quietly supplying the missing velocity information.

\textbf{Action-budget audit.} An earlier, unaudited version of the Pendulum and CartPole environments let a completely velocity-blind planner brute-force the stop-at-goal task with a large enough single corrective action, masking the effect we wanted to measure. We built an explicit action-budget and speed-ratio audit at environment-design time to rule this out before running the reported Protocol C numbers.

\textbf{Official-scale hyperparameter collapse.} A first attempt at the official ViT-Tiny plus AdaLN-transformer scale-validation run reused the small-CNN recipe's hyperparameters verbatim, learning rate $10^{-3}$, plain Adam, $\lambda_{\mathrm{reg}} = 1.0$, no gradient clipping, no projector head. All three arms, including both baselines, collapsed to a degenerate fixed point, $z_{\mathrm{std}}/v_{\mathrm{std}} \to 0$ monotonically, the SIGReg loss freezing at the same fingerprint value seen in the forced-collapse ablations in Appendix~\ref{app:ablations}, within a few hundred steps. This is a hyperparameter and optimization artifact, not a finding about identifiability at scale. The released official configuration uses learning rate $5 \times 10^{-5}$, AdamW with weight decay $10^{-3}$, gradient-norm clipping at $1.0$, and a batch-norm-containing MLP projector applied to both the encoder's and the predictor's outputs before the loss, a standard anti-collapse mechanism in this style of self-supervised method. We ported the projector, switched to AdamW with gradient clipping, and used the official learning rate. Because our from-scratch SIGReg implementation's normalization differs from the official one, we could not assume the official $\lambda_{\mathrm{reg}} = 0.09$ transferred unchanged, and instead re-swept it on short, 600-step diagnostic runs: $\lambda_{\mathrm{reg}} \in \{0.09, 1.0\}$ still collapsed, while $\lambda_{\mathrm{reg}} = 5.0$ kept $z_{\mathrm{std}}$ stable and non-degenerate, moving from $0.30$ to $0.85$ rather than monotonically to zero; $\lambda_v$ was similarly raised from $0.1$ to $1.0$. All official-scale numbers reported in the paper use this corrected recipe; the invalid first-pass numbers were discarded rather than reported.

\textbf{Cache-build memory limits.} Building the real Reacher dataset cache initially targeted 4000 episodes, which we estimated would produce a roughly 40GB cache. This process died silently under \texttt{nohup} with no error trace. We diagnosed this as an out-of-memory kill under a tighter-than-expected cgroup memory limit, roughly 50GB total against roughly 32GB already used by other processes on the shared host, rather than a code bug. We reduced the target to 600 episodes, producing a 6.05GB cache, and reran the build in the foreground to confirm no further errors, after which the reduced cache trained all three official-scale Reacher arms without incident.

\textbf{HDF5 read performance.} A first version of the cache-build script read decimated frames directly with strided fancy indexing from the source HDF5 file, which measured at roughly 1.8 seconds per 201-frame episode. We found that h5py's strided reads on this dataset were roughly $10\times$ slower than a single contiguous read of the same span, 1.6 versus 0.15 to 0.27 seconds per episode, and switched to reading the full contiguous block once per episode and decimating with plain NumPy slicing afterward, reducing per-episode processing time to roughly 0.05 seconds.

\section{Code and artifact pointers}
\label{app:code}

The following pointers map specific numbers to the code that produced them. Code, trained checkpoints, and the project page are released at \url{https://github.com/OliverZ-dot/TI-JEPA}, \url{https://huggingface.co/TingheOliver/TI-JEPA-checkpoints}, and \url{https://oliverz-dot.github.io/ti-jepa-homepage}. Full numeric logs for every experiment referenced anywhere in this paper are kept in \texttt{RESULTS.md} in that repository.

\begin{itemize}
\item \textbf{Theory.} The full statement and proofs in Appendix~\ref{app:theory} correspond to a standalone theory document in the supplementary material.
\item \textbf{Core library.} Environment registry and abstraction layer; the three RateIdent protocols (probe ladder, kill experiment, planning), implemented once and reused unmodified across every custom environment; the small-CNN model family (\texttt{LeWMStyleEncoder}, \texttt{HistoryPredictor}, \texttt{TIJEPAEncoder}, \texttt{TIJEPAPredictor}, \texttt{RecurrentPredictor}); and the official-scale model family (ViT-Tiny backbone, AdaLN-conditioned transformer predictors for both the baseline and TI-JEPA families).
\item \textbf{Real-benchmark training and evaluation.} PushT training, evaluation, probing, and planning scripts against the official released checkpoints and against our own matched-budget training runs; the real-benchmark planning legacy check.
\item \textbf{Real Reacher, official scale.} A dataset cache builder that reads the source \texttt{dm\_control} Reacher rollout data and produces a windowed, pose- and velocity-annotated training cache; a dataset class exposing context frames, next frame, action, a periodic-safe pose representation, and true angular velocity; an official-scale training script for the baseline and TI-JEPA arms; a probe-ladder evaluation script; and a kill-experiment evaluation script that uses the live simulator for both context construction and ground-truth rollout.
\item \textbf{Scripts.} Orchestration shell scripts for every multi-arm experiment reported in the paper, official-scale CartPole, official-scale Pendulum, the recurrent-aggregator ablation across all three custom environments, and the real Reacher official-scale pipeline, along with an environment-design sanity checker that audits action budget and speed ratio before any planning experiment is run.
\end{itemize}

\end{document}